%% file: main.tex
\documentclass[11pt]{article}

\usepackage[margin=1in]{geometry}
\usepackage{times}
\usepackage[round,authoryear]{natbib}
\input{math_commands.tex}

\usepackage{titletoc}
\usepackage{amsmath,amssymb,amsthm}
\usepackage{booktabs}
\usepackage{multirow}
\usepackage{graphicx}
\usepackage{algorithm}
\usepackage{algpseudocode}
\usepackage{microtype}
\usepackage{float}
\usepackage{placeins}
\usepackage{xcolor}
\usepackage{hyperref}
\usepackage{url}
\hypersetup{
  pdftitle={Terminal Symmetry as a Carrier of Asymmetric Process Knowledge: Statewise Refinement for Anytime Verified Construction},
  pdfauthor={Yi Liu}
}
\newtheorem{theorem}{Theorem}
\newtheorem{proposition}{Proposition}
\newtheorem{lemma}{Lemma}

\newcommand{\Top}{\operatorname{Top}}
\newcommand{\AUC}{\operatorname{AUC}}
\newcommand{\ind}{\mathbb{I}}

\title{Terminal Symmetry as a Carrier of\\Asymmetric Process Knowledge:\\Statewise Refinement for Anytime Verified\\Construction}

\author{Yi Liu\\
\normalsize MS Student\\
\normalsize University of Science and Technology of China\\
\normalsize \texttt{scnuliuyi@mail.ustc.edu.cn}}
\date{}

\begin{document}

\maketitle

\begin{abstract}
Many sequential construction tasks have exact terminal symmetries even though execution is directed and depends on history. Process evidence supplies order. Terminal correspondence transports it between equivalent outcomes; the realized state updates relevance. These roles define a carrier framework: transport what the outcome preserves; refine what history changes. SymBuild combines transported process and state residual ranks by ordinal rank meet. Its rank threshold sets equal the corresponding unions, yielding a tight worst-case verifier query bound under prefix information. We evaluate SymBuild in three construction domains: computer-aided design (CAD) assembly, Mini-Programs, and exact-fill packing, and its ordinal adaptation in target removal. SymBuild improves the area under the anytime verified success curve by up to 6.77, 21.75, and 8.68 points over Static in the three construction domains. Refresh gains recur beyond SymBuild under alternative aggregation, planning, and learned scoring methods; on Geometric Reasoning Network (GRN) target removal, direct Combined refresh has the lowest mean verifier query score at all three scales and reduces distinct scored states by factors of 6.48--12.22 relative to refreshed population-based search. Together, these results support the carrier framework and demonstrate that SymBuild is an effective, analyzable method for anytime verified construction.
\end{abstract}

\section{Introduction}

Many sequential tasks have exact terminal symmetries even though execution is directed and depends on history.  A chair may be unchanged by exchanging legs, a program output by renaming, and an exact-fill packing by relabeling bins, yet blockers, dependencies, and remaining capacity determine which next actions are useful.  Terminal equivalence offers a distinct reuse opportunity: process evidence from one terminal frame can guide an equivalent frame, while accepted transitions change its relevance.  Yet terminal symmetry equates outcomes, not the processes that realize them.  With costly verification, this distinction determines where limited verifier effort is spent.  This raises a basic question: \emph{what decision information can exact terminal symmetry provide when the realizing process itself is asymmetric?}

The \textbf{carrier framework} assigns these roles explicitly: process evidence supplies directionality, terminal correspondence transports it into an equivalent target frame, the realized state updates which parts of the transported structure remain decision-relevant after each transition, and a fixed verifier certifies execution.  Different proposal methods can use the same transported structure and current-state evidence.

Prior work exploits symmetry in planning search \citep{fox1999symmetry,fox2005action}, assembly reasoning \citep{liu1990assembly}, and equivariant state-action models, trajectories, and canonical representations \citep{vanderpol2020mdp,zhao2023symplan,tie2025etseed,kaba2023canonicalization}.  We study exact symmetry only at terminal outcomes: process evidence supplies directed structure \citep{seminara2024taskgraph}, while terminal correspondence transports it across equivalent outcomes.  Theorem~\ref{thm:nonident} makes the limit explicit: even the same symmetric terminal object and observed demonstration can be consistent with different commutativity and precedence relations.

\begin{figure}[t]
  \centering
  \includegraphics[width=.97\linewidth]{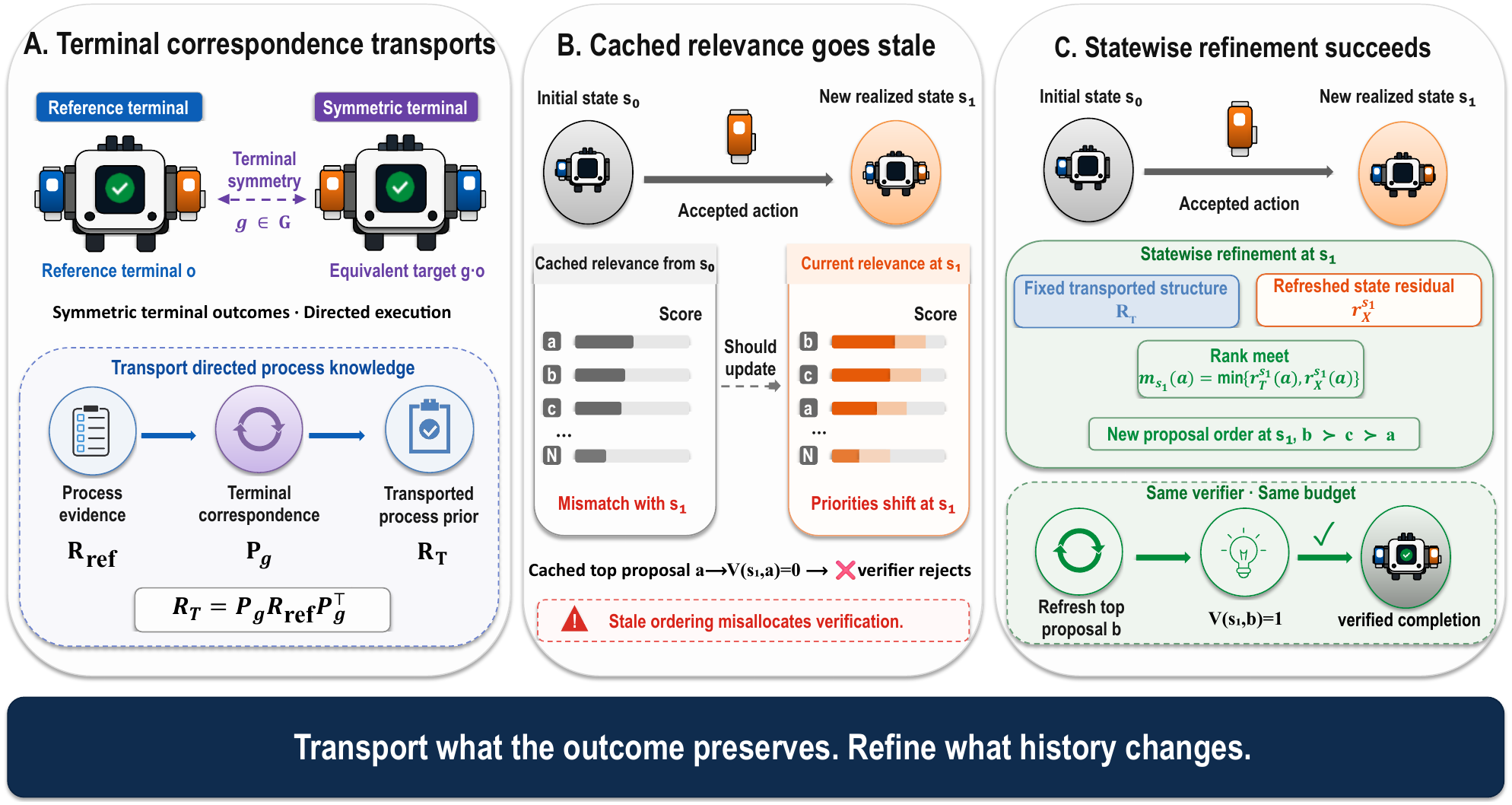}
  \caption{\textbf{Terminal correspondence transports directed process knowledge; realized state refines its decision value after accepted transitions.}
  (A) Process evidence grounds an asymmetric prior and a terminal automorphism transports it into the target frame.
  (B) After the shared first action changes feasibility, a cached composition becomes stale and the verifier rejects its next proposal.
  (C) \textup{SymBuild} refreshes the state residual and reaches a verified plan under the same budget and verifier.}
  \label{fig:thesis}
\end{figure}

This yields a simple rule: transport what the outcome preserves; refine what history changes.  A transported process prior carries asymmetric order between equivalent terminal frames; a state residual tracks accessibility, blockage, satisfied dependencies, and remaining capacity.  \textup{SymBuild} instantiates the framework with ordinal rank meet.  At each rank threshold, rank meet selects the union of the two input sets, making verifier allocation analyzable and yielding the query guarantee in Theorem~\ref{thm:complete}.  Figure~\ref{fig:thesis} illustrates how refresh redirects verification after a shared first action. The decomposition yields two empirical signatures: disrupting terminal correspondence should weaken transported process value, while suppressing post-transition refresh should misallocate later verification. Empirically, SymBuild improves anytime verified success under matched budgets; other instantiations of the framework also yield performance gains in multiple domains.

Our contributions are:
\begin{itemize}
  \item \textbf{Carrier framework.} We formulate symmetric terminal outcomes with asymmetric processes as information flow: process evidence supplies directed structure, terminal correspondence transports it, the realized state updates the decision relevance of the transported structure, and a fixed verifier certifies execution.
  \item \textbf{SymBuild and theory.} \textup{SymBuild} uses ordinal rank meet.  Its rank threshold sets equal the unions of the two input sets, yielding the completion guarantee of Theorem~\ref{thm:complete} and a tight worst-case verifier query bound; the framework also predicts that refresh and static diverge after a shared first accepted action.
  \item \textbf{Evidence in multiple domains and matched evaluation.} Our protocol separates transport and refresh effects. \textup{SymBuild} improves anytime verified efficiency in three construction domains, with AUC gains up to 21.75 percentage points over Static. Framework instantiations show refresh gains in four domains under alternative aggregation, planning, and learned scoring. On GRN target removal \citep{bouhsain2025grn}, direct Combined refresh has the lowest query score at all scales, using 6.48--12.22 times as few distinct scored states.
\end{itemize}

\section{Related Work}

\paragraph{Structure from symmetry and terminal outcomes.}
Equivariant RL and differentiable planning encode symmetry in state-action values, policies, operators, and action sampling \citep{vanderpol2020mdp,wang2022eqq,zhao2023symplan,nguyen2023partial,zhao2024equivariant,klee2026raven}; goal-conditioned robotics separates task and motion symmetries \citep{mittal2024tasksymmetry}, while experience, trajectory, and generative methods reuse symmetry across trajectories, structures, and policy spaces \citep{lin2020iter,brehmer2023edgi,tie2025etseed,zhao2024pard,kim2025sagfn,hadjiloizou2026crossspace}.  Goal-conditioned robot policies map desired configurations or images to actions \citep{seita2021goal,ghosh2024octo}; visual planning and world model methods use intermediate plans or generated goals for long-horizon control \citep{ebert2018robustness,yang2025vis2plan,zhou2026act2goal}.

\paragraph{Symmetry in planning, solving, and assembly.}
Classical planning uses exact or approximate symmetry to reduce redundant search or guide action ordering \citep{fox1999symmetry,pochter2011symmetries,fox2005action}.  Dynamic SAT solving transports logical consequences under formula symmetry: symmetry propagation derives implications, while symmetric explanation learning transports explanation clauses and selects images that restrict the current partial assignment \citep{devriendt2012symprop,devriendt2017sel}.  Our framework transports process evidence into proposal information, with acceptance determined by the verifier.  Assembly planning uses component symmetries and group theory to simplify geometric and kinematic reasoning \citep{liu1990assembly,popplestone1990group}.  Across these settings, symmetry structures search, inference, or reuse at the level where equivalence is available.

\paragraph{Representative selection and imperfect symmetry.}
Canonicalization selects orbit representatives \citep{kaba2023canonicalization,lin2026adaptivecanon,zhou2026canonical}, while soft, approximate, probabilistic, and partial methods modulate equivariance when symmetry is imperfect \citep{finzi2021rpp,wang2022approx,xie2024breaking,wang2024relaxed,lawrence2025probabilistic,chang2026partially}.  Related work also studies how symmetry interacts with evolving search states and equivariant solution families. Appendix~\ref{app:recent-symmetry} compares these operational roles \citep{devriendt2012symprop,devriendt2017sel,lee2023emmp,chang2026partially}.

\paragraph{Search under expensive evaluation.}
Partial-order planning \citep{mcallester1991systematic} and partial-order reduction \citep{xu2011por} exploit ordering structure to constrain or reduce planning search; anytime heuristic planning and search \citep{likhachev2003ara,hansen2007anytime,richter2010lama} improve solutions as computation continues, multi-heuristic search \citep{aine2016multi} combines multiple heuristics, and LazySP \citep{dellin2016lazy} targets shortest-path search with expensive edge evaluations. SayCan and VeriGraph update execution decisions from the realized state \citep{ichter2023saycan,ekpo2026verigraph}.  These lines motivate two complementary roles for guidance: reusable structure and state-dependent decision value.  The carrier view brings them together by transporting independently grounded process structure through terminal correspondence, refining its relevance in the realized state, and leaving acceptance to a fixed verifier. Section~\ref{sec:search-controls} tests this decomposition across search interfaces.
\section{Terminal Symmetry as a Carrier}

\subsection{Terminal Correspondence as Transport}
\label{sec:formal}

Process evidence supplies directed process structure, and terminal symmetry provides a transport map between equivalent outcomes.

\subsubsection{Construction systems and terminal lifts}

A discrete construction system is $\mathcal C=(\mathcal S,\mathcal A,F,s_0,\Psi)$, where $F:\mathcal S\times\mathcal A\rightharpoonup\mathcal S$ is a deterministic partial transition and $\Psi$ maps a final state to a semantic object.  A complete executable trajectory is $\tau=(a_1,\ldots,a_T)\in\mathcal T_{\mathcal C}$ and its terminal object is $O_{\mathcal C}(\tau)=\Psi(s_T)$.

Let a group $G$ act on terminal objects.  A potentially partial and set-valued lift $\mathsf L_g(\tau)$ satisfies
\begin{equation}
  O_{\mathcal C}(\tau')=g\cdot O_{\mathcal C}(\tau)
  \qquad \forall\tau'\in\mathsf L_g(\tau).
  \label{eq:terminal-covariance}
\end{equation}
\Eqref{eq:terminal-covariance} specifies terminal covariance.  Symmetry is defined in this outcome model; experiments compute or estimate its correspondence from domain observations.  On corresponding entity or event indices, $P_g$ is the selected permutation matrix.  Independently obtained process evidence $R_{\rm ref}$ defines transported pairwise relations by
\begin{equation}
  R_T:=P_g R_{\rm ref}P_g^\top.
  \label{eq:relation-transport}
\end{equation}
Conjugation reindexes precedence, commutativity, or event correspondence.  For a relation $R$ in one frame, $R=P_gRP_g^\top$ means invariance under that permutation.  Domain maps additionally transform directions or assignment labels before scoring candidate actions (Appendix~\ref{app:setup}).

\subsubsection{What terminal symmetry does not determine}

Process evidence supplies relation content; terminal correspondence determines its indices in the target frame.

\begin{theorem}[Terminal symmetry does not determine process relations]
\label{thm:nonident}
Over finite construction systems with deterministic partial transitions, no estimator that observes only a terminal object and its stabilizer can universally recover true precedence or commutativity.  The result also covers one observed demonstration whose sampling policy may omit counterfactual action orders.  On a witness class containing two systems, every randomized estimator has worst-case error at least $1/2$.
\end{theorem}

A commuting diamond and a forced chain share the same terminal object, stabilizer, and demonstration, illustrating why directed relations require independent process evidence (Appendix~\ref{app:proofs}).

\subsection{Carrier Framework and Statewise Refinement}
\label{sec:method}

\subsubsection{Three information roles, two timescales, one verifier}

Process evidence and terminal correspondence yield a transported structure $R_T$ that is fixed for the episode; restricting it to candidates $A_s$ induces the transported process rank $r_T^s:A_s\to[1,|A_s|]$.  Evidence from the realized state supplies a state residual rank $r_X^s$, refreshed after accepted actions as geometry, capacity, dependencies, or extension viability change; this recomputation realizes statewise refinement.  A fixed verifier $V_s(a)$ certifies execution.

\begin{figure}[H]
  \centering
  \includegraphics[width=.97\linewidth]{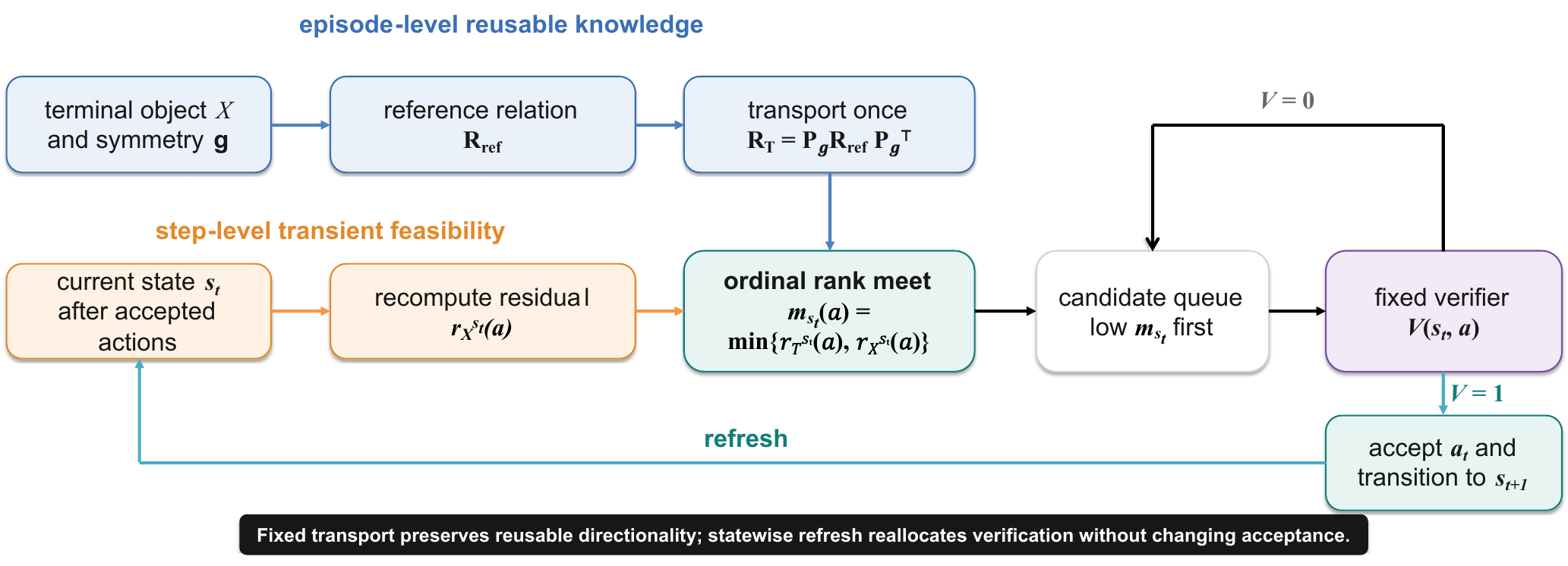}
  \caption{\textbf{Mechanism of \textup{SymBuild}.}  Process evidence grounds the directed prior; terminal correspondence transports it once.  Process ranks use the remaining candidates; state evidence refreshes the residual after accepted transitions.  Their ordinal meet orders proposals to a fixed verifier.}
  \label{fig:mechanism}
\end{figure}

The framework specifies information roles; proposal ordering remains modular.  \textup{SymBuild} uses ordinal rank meet so a proposal remains early whenever either rank places it early:
\begin{equation}
  m_s(a)=\min\{r_T^s(a),r_X^s(a)\}.
  \label{eq:rankmeet}
\end{equation}
Lemma~\ref{lem:union} gives the rank threshold identity.  Figure~\ref{fig:mechanism} and Algorithm~\ref{alg:symbuild} (Appendix~\ref{app:complete-tables}) summarize the planner.  \textup{SymBuild} orders actions by $(m_s(a),\kappa_s(a))$, with a deterministic secondary key, and accepts the first action approved by $V_s$.  After acceptance, it recomputes current ranks while retaining the transported structure. Static retains initial residual information.  Appendix~\ref{app:setup} specifies domain rank updates and sorting keys.  The same ranks can guide separate proposal queues, tested in Section~\ref{sec:search-controls} with scheduling detailed in Appendix~\ref{app:search-controls}.

\FloatBarrier

\paragraph{Instantiations.}
CAD transports blocker order and refreshes accessibility; Mini-Programs combine learned and observed relations and update remaining dependencies; packing transports a reference assignment and refreshes capacity-dependent viability (Appendix~\ref{app:setup}).  Rank meet realizes the highest rank combination rule \citep{ho1994decision} with Lemma~\ref{lem:union}'s union identity. Section~\ref{sec:search-controls} tests alternative aggregation and planning methods.

\subsubsection{Prefix coverage and query optimality}

For a weak rank $r$, its rank threshold set is $\Top_k(r)=\{a:r(a)\le k\}$, with size $B_r(k)=|\Top_k(r)|$.

\begin{lemma}[Union coverage]
\label{lem:union}
For the rank meet in \eqref{eq:rankmeet},
\begin{equation}
 \Top_k(m_s)=\Top_k(r_T^s)\cup\Top_k(r_X^s).
\end{equation}
If either expert's rank threshold set contains an accepted action, rank meet finds one in at most $B_T^s(k)+B_X^s(k)$ verifier calls, or $2k$ for strict ranks and integer $1\le k\le|A_s|$.
\end{lemma}

\begin{theorem}[Optimality under prefix information and completion guarantee]
\label{thm:complete}
Let $U_s(k)=\Top_k(r_T^s)\cup\Top_k(r_X^s)$.  Suppose every sequence of accepted transitions from $s_0$ has length at most $H$ and every candidate set is finite.  Assume $V_s(a)=1$ only if $F(s,a)$ admits a complete suffix, and every incomplete reachable state contains an action $a\in U_s(k_s)$ with $V_s(a)=1$.  With sufficient query budget, \textup{SymBuild} completes in $T\le H$ accepted transitions using at most
\begin{equation}
  \sum_{t=0}^{T-1}|U_{s_t}(k_{s_t})|
  \le \sum_{t=0}^{T-1}\bigl(B_T^{s_t}(k_{s_t})+B_X^{s_t}(k_{s_t})\bigr)
\end{equation}
queries.  At one state, meet attains the tight worst-case bound $|U_s(k)|$ among deterministic query policies when the only promise is a feasible action in $U_s(k)$.
\end{theorem}

Union coverage supplies an accepted action, and extension preservation allows this argument to continue to completion.  Appendix~\ref{app:completion-proof} proves the guarantees and details finite-budget execution and completeness over the full candidate graph.

\subsection{Anytime Value Under Expensive Verification}
\label{sec:anytime}

Statewise refinement reallocates expensive verification, so performance is naturally measured by completion before an uncertain resource limit.  Let $Q_\pi$ be the verifier resource required to reach a complete verified solution and $F_\pi(b)=\Pr(Q_\pi\le b)$.  For a predeclared budget distribution $\mu$,
\begin{equation}
  \AUC_\mu(\pi)=\sum_{b\in\mathcal B}\mu(b)F_\pi(b)=\Pr_{\omega,B}(Q_\pi(\omega)\le B),\qquad B\sim\mu.
\end{equation}
Thus anytime AUC is expected success under random interruption. Appendix~\ref{app:proofs} gives the formal derivation and the capped cost identity.

A shared first action can leave Static testing an infeasible proposal, while refresh selects a feasible one and completes within two calls using the same prior and fixed verifier.  Proposition~\ref{prop:strict} in Appendix~\ref{app:refresh-separation} gives the construction and proof. Figure~\ref{fig:refresh-diagnostic} tests this pattern empirically.

\section{Experiments}
\label{sec:experiments}

Experiments evaluate the carrier framework in four domains.  \textup{SymBuild} receives performance comparisons, ablations, and mechanism analyses in CAD assembly, Mini-Programs, and exact-fill packing, with its ordinal adaptation evaluated in target removal.  Additional framework tests use alternative aggregation in the construction domains, planning variants in Mini-Programs, and learned GRN scoring with direct ranking and population-based search.

\subsection{Domains and protocol}

We evaluate CAD assembly (72 assemblies in the call panel and 48 in the independent work panel), 480 Mini-Programs, 480 symmetric exact-fill packings, and 1,135 solvable target removal episodes from official GRN out-of-distribution (OOD) scenes.  ID denotes in-distribution settings and pp denotes percentage points.  Static omits post-transition proposal refresh while accepted actions update candidates and required planner state.  Matched pairs share instances, observables, candidates, verifier, and budgets.  Primary construction pairs and their post-transition diagnostics also share first actions.  GRN and CDGS use five seeds, one frozen GRN checkpoint, and a $5N$ cap.  Direct ranking pairs also share first queried actions.  Appendix~\ref{app:setup} specifies domain-specific seeds, construction, and inference.  Table~\ref{tab:protocol} summarizes the three \textup{SymBuild} domains.  Appendix~\ref{app:search-controls} gives the planning controls, and Appendix~\ref{app:grn-search} adapts Compositional Diffusion with Guided Search (CDGS, \citealp{mishra2026cdgs}).  Appendix Table~\ref{tab:open-code-compare} identifies the related methods and their roles.

\subsection{\textup{SymBuild} in Three Construction Domains}

\begin{figure}[t]
  \centering
  \includegraphics[width=\linewidth]{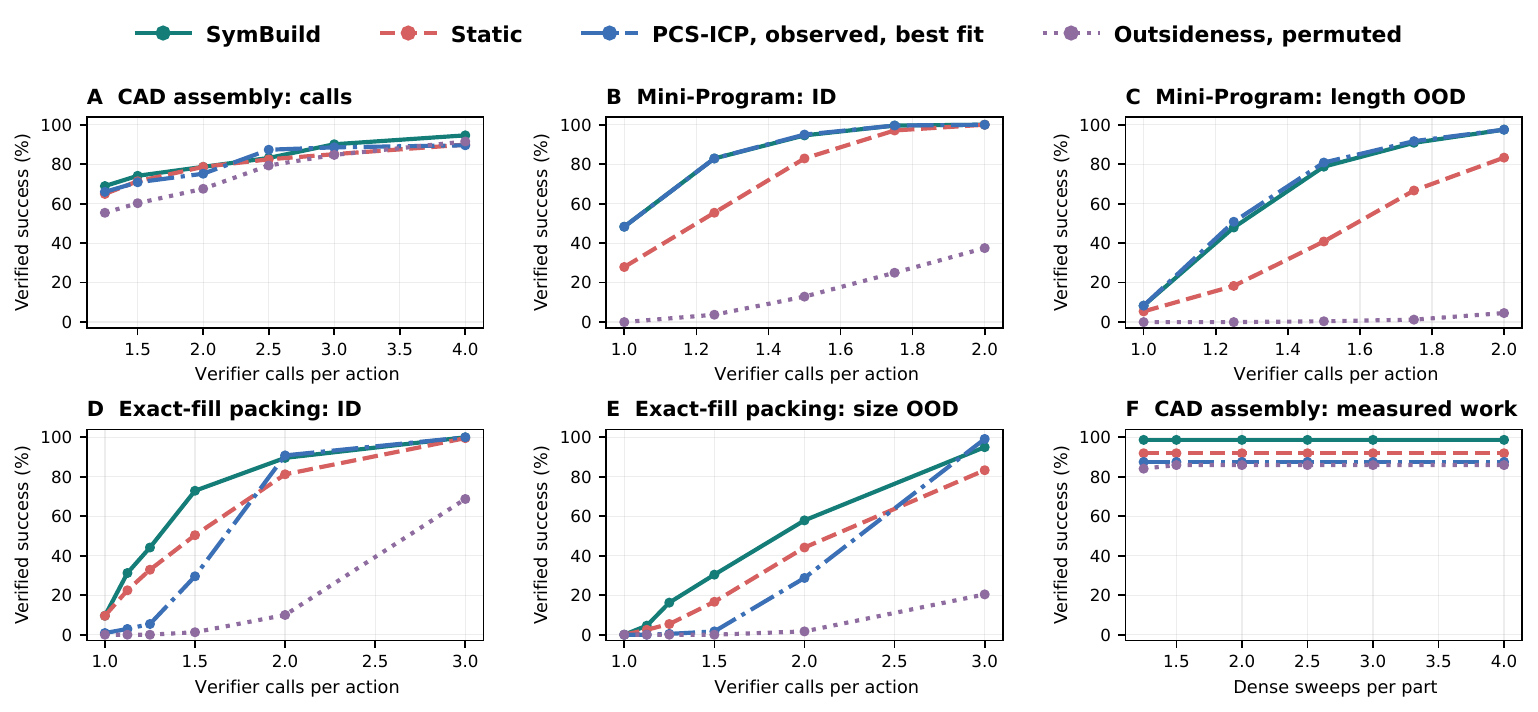}
  \caption{\textbf{Statewise refinement shifts the frontier of verified success throughout the predeclared budget range.}  Other single-channel and permuted controls appear where defined; the flat CAD work curves show that the gain comes from additional verified completions.}
  \label{fig:frontiers}
\end{figure}

In CAD assembly, Mini-Programs, and exact-fill packing, \textup{SymBuild} improves verified efficiency over Static in all six aggregates (Table~\ref{tab:main}).  With transported process structure and verifier fixed, \textup{SymBuild} refreshes state relevance after accepted transitions, while Static retains its initial relevance ranking.  Figure~\ref{fig:frontiers} also includes PCS-ICP, a CAD control using trimmed iterative closest point (ICP) matching \citep{besl1992method,chetverikov2002trimmed} in the process coordinate system (PCS).

The same ordering holds across the predeclared budgets, with separations at intermediate budgets in CAD calls, Mini-Programs, and packing (Figure~\ref{fig:frontiers}).  CAD work has reached its plateau. Appendix~\ref{app:frontiers} examines smaller budgets.  Pointwise ordering preserves the result under nonnegative budget weighting, with strict gains whenever separating budgets receive positive mass (Appendix~\ref{app:proofs}).  Appendix Table~\ref{tab:all-frontiers} gives exact values. Appendix Table~\ref{tab:replication} reports subgroup replications.

\begin{table}[h]
\centering
\caption{\textbf{\textup{SymBuild} in three construction domains.}  Matched pairs share channels, candidates, first actions, and verifier. Static omits post-transition refresh.  AUC entries are percentages with paired 95\% confidence intervals (CIs).  Cost reduction is the paired decrease in capped normalized verifier cost, with unsuccessful runs charged the cap.}
\label{tab:main}
\scriptsize
\renewcommand{\arraystretch}{0.90}
\setlength{\tabcolsep}{3.4pt}
\begin{tabular}{llrrrrr}
\toprule
Domain & Setting & $N$ & Static & \textup{SymBuild} & $\Delta$AUC $[95\%\,\mathrm{CI}]$ & Cost reduction\\
\midrule
CAD & ID (calls) & 72 & \underline{78.73} & \textbf{81.60} & $+2.86$ $[.12,5.76]$ & .0253\\
CAD & ID (work) & 48 & \underline{91.93} & \textbf{98.70} & $+6.77$ $[1.30,14.06]$ & .0523\\
Program & ID (calls) & 240 & \underline{72.67} & \textbf{85.08} & $+12.42$ $[9.50,15.33]$ & .0737\\
Program & length OOD & 240 & \underline{42.92} & \textbf{64.67} & $+21.75$ $[18.33,24.92]$ & .1351\\
Packing & ID (calls) & 240 & \underline{49.38} & \textbf{57.92} & $+8.54$ $[7.01,10.14]$ & .0539\\
Packing & size OOD & 240 & \underline{25.35} & \textbf{34.03} & $+8.68$ $[7.01,10.42]$ & .0981\\
\bottomrule
\end{tabular}
\end{table}

\subsection{Post-Transition Effect of Statewise Refinement}

\begin{figure}[t]
  \centering
  \includegraphics[width=.84\linewidth]{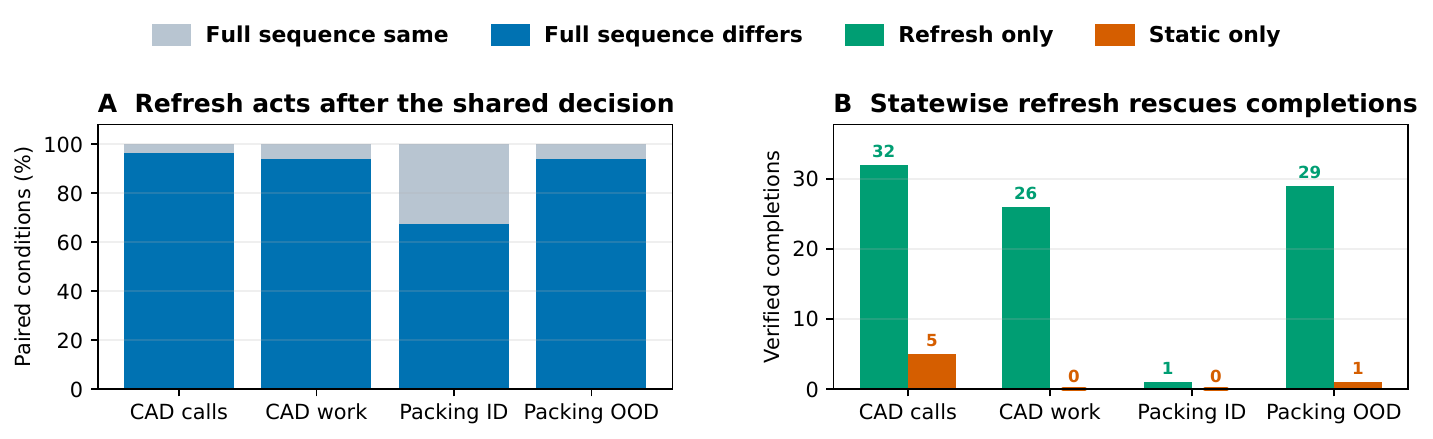}
  \caption{\textbf{Statewise refresh acts after the shared first decision.}  Accepted sequences diverge after the transition, and verified completions unique to the refreshed policy dominate.}
  \label{fig:refresh-diagnostic}
\end{figure}

Proposition~\ref{prop:strict} predicts that refreshed and static policies share the first accepted action and diverge after the realized transition changes residual relevance.  The primary rank meet comparison shows this pattern directly: \textup{SymBuild} and Static agree on the first accepted action in the paired CAD and packing panels, then diverge, with more verified completions unique to \textup{SymBuild} (Figure~\ref{fig:refresh-diagnostic}).  An independent CAD panel reproduces the post-transition effect (Table~\ref{tab:ablation}(b)).

\subsection{Information Roles and Robustness in \textup{SymBuild}}
\label{sec:boundary-main}

The ablations separate the two information roles.  CAD assembly and exact-fill packing are more sensitive to transport; state evidence dominates in Mini-Programs.  In the independent CAD panels, Algorithm compares \textup{SymBuild} with PCS-ICP and an outsideness control adapted from ASAP~\citep{tian2024asap}, Timing compares \textup{SymBuild} with Static, and Alignment compares aligned with permuted transport; all share the verifier.  Observed process topology adds complementary mechanism evidence (Table~\ref{tab:ablation}).  Here, IKEA Manuals at Work~\citep{liu2024ikea} provides real furniture-assembly videos aligned with parts, manuals, and substeps. Appendix~\ref{app:setup} gives the protocol and definitions.  Transport contributes most when correspondence relocates physical constraints; state evidence can dominate when the residual exposes current precedence.  Complete channel comparisons are in Appendix~\ref{app:complete-tables}.

\begin{table}[!hb]
\centering
\caption{\textbf{Evidence for distinct information roles.}  (a) AUC (\%) ablations; (b) CAD mechanism point estimates.  In (b), Algorithm reports two comparisons, first with PCS-ICP and then outsideness.  Timing lists final success for \textup{SymBuild} then Static. Alignment lists aligned then permuted transport.  In each comparison, success gain is the first condition minus the comparator, and call saving is the comparator minus the first condition.  Positive values therefore favor the first condition.  (c) Gains in Top-1 accuracy for predicting the observed next action (pp), with paired 95\% CIs.}
\label{tab:ablation}
\scriptsize
\renewcommand{\arraystretch}{0.74}
\setlength{\tabcolsep}{3.1pt}
\begin{tabular*}{\textwidth}{@{\extracolsep{\fill}}llrrrrr@{}}
\toprule
\multicolumn{7}{l}{\textbf{(a) Refresh and information channel ablations} (AUC $\uparrow$)}\\
Domain & Panel & $N$ & \textup{SymBuild} & Static & Process-only & State-only\\
\midrule
CAD & calls & 72 & 81.60 & 78.73 & 79.57 & 73.12\\
CAD & work & 48 & 98.70 & 91.93 & 87.50 & 85.63\\
Program & ID & 240 & 85.08 & 72.67 & 84.83 & 85.17\\
Program & length OOD & 240 & 64.67 & 42.92 & 63.58 & 65.83\\
Packing & ID & 240 & 57.92 & 49.38 & 50.21 & 38.26\\
Packing & size OOD & 240 & 34.03 & 25.35 & 28.40 & 21.67\\
\midrule
\end{tabular*}

\begin{tabular*}{\textwidth}{@{\extracolsep{\fill}}llrlrr@{}}
\multicolumn{6}{l}{\textbf{(b) Independent CAD mechanism contrasts}}\\
Test & Panel & Final success (\% $\uparrow$) & Comparator & Success gain (pp $\uparrow$) & Call saving (calls per part $\uparrow$)\\
\midrule
\multirow{2}{*}{Algorithm} & \multirow{2}{*}{120 objects $\times$ 4 lifts} & \multirow{2}{*}{84.58} & PCS-ICP & $+5.00$ & $+.143$\\
& & & outsideness & $+7.71$ & $+.168$\\
Timing & 120 objects $\times$ 4 lifts & $84.38\,/\,80.00$ & Static & $+4.38$ & $+.152$\\
Alignment & 60 objects $\times$ 4 lifts & $81.25\,/\,65.00$ & permuted transport & $+16.25$ & $+.637$\\
\midrule
\end{tabular*}

\begin{tabular*}{\textwidth}{@{\extracolsep{\fill}}llrrrrr@{}}
\multicolumn{7}{l}{\textbf{(c) Evidence from observed process topology} (Top-1 accuracy gain $\uparrow$)}\\
Evidence & Panel & \multicolumn{2}{c}{Process structure} & \multicolumn{2}{c}{Action core} & Hybrid\\
\midrule
IKEA topology & 36 families, 97 videos, 1,022 states & \multicolumn{2}{c}{$+14.14$ $[6.41,20.18]$} & \multicolumn{2}{c}{$+19.20$ $[10.26,25.56]$} & $+2.64$ $[.77,4.38]$\\
\bottomrule
\end{tabular*}

\end{table}

Beyond the tabulated contrasts, the Timing panel shows mean verifier calls per part of $1.569$ for \textup{SymBuild} and $1.721$ for Static, with $96.46\%$ sequence disagreement.  In (c), Process structure uses states with terminal ambiguity, Action core restricts to cases where the realized prefix breaks every relevant transport, and Hybrid uses the section model only on ambiguous states.  Appendix Table~\ref{tab:cad-mechanism-ci} gives the CAD CIs.

Correspondence and certification tests show two robustness effects.  Transport value decreases as correspondence mismatch increases, while \textup{SymBuild} retains positive true AUC gains over Static when certification is approximate (Table~\ref{tab:boundary-main}).

\begin{table}[H]
\centering
\caption{\textbf{Correspondence fidelity and practical verification value.}  (a) Each cell gives AUC first and final verified success second, both in percent, under graded deterministic mismatch of action identities; Appendix~\ref{app:robustness} gives construction details.  (b) At 5\% verifier noise, $\Delta$AUC is \textup{SymBuild} minus Static; FC denotes false certification, meaning completion without true success.}
\label{tab:boundary-main}
\scriptsize
\renewcommand{\arraystretch}{0.82}
\setlength{\tabcolsep}{3.3pt}
\begin{tabular}{lrrrrr}
\toprule
\multicolumn{6}{l}{\textbf{(a) Correspondence fidelity controls transport value} (AUC and success $\uparrow$)}\\
Requested mismatch & $0\%$ & $10\%$ & $25\%$ & $50\%$ & $100\%$\\
\midrule
Program ID & \textbf{85.08/100.00} & 75.33/98.33 & 63.42/95.42 & 40.58/77.92 & 13.58/36.67\\
Program length OOD & \textbf{64.67/97.50} & 49.58/88.75 & 32.50/70.00 & 11.42/31.67 & .75/2.92\\
\midrule
\multicolumn{6}{l}{\textbf{(b) Approximate verification preserves refresh value}}\\
Condition & ID $\Delta$AUC $\uparrow$ & Length OOD $\Delta$AUC $\uparrow$ & \multicolumn{3}{l}{Reliability marker}\\
\midrule
5\% false positive & $+17.47$ $[13.70,21.25]$ & $+20.48$ $[17.18,23.76]$ & \multicolumn{3}{l}{FC reduction $\uparrow$: $10.89$ pp ID; $14.00$ pp OOD}\\
5\% false negative & $+8.65$ $[6.25,11.13]$ & $+15.01$ $[12.36,17.75]$ & \multicolumn{3}{l}{No false certification on either split}\\
\bottomrule
\end{tabular}
\end{table}

These interventions trace both parts of the information flow: transport value falls with correspondence mismatch, while refresh gains persist throughout the $1$--$10\%$ noise sweep and false certification is lower under false positive noise (Appendix Table~\ref{tab:verifier-noise}).

\subsection{Framework Tests Beyond \textup{SymBuild}: Aggregation and Planning}
\label{sec:search-controls}

The Mini-Program planning controls keep the 480 objects, candidates, verifier, and budgets fixed.  Table~\ref{tab:search-main} compares alternative planners and tests refresh within different planning mechanisms.  \textup{SymBuild} outperforms the canonical slice and weighted A* with LazySP-style replanning on both splits, demonstrating more effective proposal ordering from shared process and state information under limited verification budgets.

\begin{table}[H]
\centering
\caption{\textbf{Comparison under alternative planning methods.}  AUC gaps with paired 95\% CIs. Appendix Table~\ref{tab:search-full} adds final success, selector time, and search expansions.}
\label{tab:search-main}
\scriptsize
\renewcommand{\arraystretch}{0.80}
\setlength{\tabcolsep}{3.4pt}
\resizebox{\textwidth}{!}{%
\begin{tabular}{lrrr}
\toprule
Comparison & ID gap $\uparrow$ & Length OOD gap $\uparrow$ & Interface feature\\
\midrule
Difference between \textup{SymBuild} and canonical slice with static meet & $+12.58$ $[9.67,15.50]$ & $+21.75$ $[18.42,25.17]$ & canonical representative\\
Difference between \textup{SymBuild} and weighted A* with LazySP-style replanning & $+4.42$ $[2.58,6.33]$ & $+4.25$ $[1.83,6.67]$ & lazy replanning\\
Difference between POCL-style refresh and fixed & $+1.08$ $[.25,2.00]$ & $+3.83$ $[2.58,5.08]$ & partial-order agenda\\
Difference between MHA*-style refresh and fixed & $-.42$ $[-1.75,.83]$ & $+5.00$ $[2.83,7.17]$ & scheduling with two queues\\
\bottomrule
\end{tabular}}
\end{table}

Within partial-order causal-link (POCL)-style \citep{mcallester1991systematic} and MHA*-style \citep{aine2016multi} planning, refresh raises length OOD AUC by $3.83$ and $5.00$ pp, respectively.  These paired gains extend the framework to partial-order agendas and scheduling with two queues. Appendix Table~\ref{tab:search-full} reports completion and computational costs.  Alternative aggregation provides complementary evidence: the SayCanPay-adapted product rule \citep{hazra2024saycanpay} retains refresh gains in all six construction panels after the shared first action, while a weighted sum selected on development data strengthens Mini-Program results (Appendix Tables~\ref{tab:refresh-table} and~\ref{tab:aggregation-full}).  Rank meet is the \textup{SymBuild} choice because it outperforms Borda count \citep{ho1994decision} and rank join on the CAD development panel and supplies the rank threshold identity in Lemma~\ref{lem:union} (Appendix~\ref{app:aggregation-transfer}).

\subsection{Learned Scoring and CDGS-style Search in GRN Target Removal}

Following transport and refinement tests in the construction domains, GRN target removal evaluates SymBuild-style ordinal fusion and refresh with identity correspondence on scene objects.  Combined sums learned GRN feasibility and annotated process scores.  Rank meet fuses their ordinal ranks.  Direct ranking selects individual actions, while CDGS-style search evaluates short imagined sequences.  Static caches initial raw scores, and refresh recomputes them after accepted removals.  Appendix~\ref{app:grn-search} gives the full paired protocol.

\begin{table}[t]
\centering
\caption{\textbf{GRN scoring and population-based search inspired by CDGS.} Panels (a) and (b) test refresh against matched Static controls using mean verifier query scores ($\downarrow$). Failures receive $5N+1$ under a $5N$ cap. Each $\Delta$ is refresh minus Static for the same method with a paired 95\% CI. Negative values favor refresh. In (a), bold and underlined means mark the best and second-best scores per scale. Panel (b) repeats direct Combined refresh for comparison. Panel (c) reports distinct scored GRN states and sampled sequence positions per target. The reduction factor divides CDGS--refresh state counts by direct Combined refresh state counts.}
\label{tab:cdgs-main}
\scriptsize
\renewcommand{\arraystretch}{0.70}
\setlength{\tabcolsep}{2.7pt}
\resizebox{\textwidth}{!}{%
\begin{tabular}{lrrrrrl}
\toprule
\multicolumn{7}{l}{\textbf{(a) Direct action ranking}}\\
Split & Targets & \multicolumn{1}{l}{Method} & Static mean & Refresh mean & \multicolumn{2}{l}{$\Delta$ $[95\%\,\mathrm{CI}]$}\\
\midrule
OOD10 & 129 & \multicolumn{1}{l}{GRN} & 5.457 & 5.132 & \multicolumn{2}{l}{$-0.326\;[-0.743,0.066]$}\\
OOD10 & 129 & \multicolumn{1}{l}{Rank meet} & 5.352 & 4.747 & \multicolumn{2}{l}{$-0.605\;[-1.433,-0.025]$}\\
OOD10 & 129 & \multicolumn{1}{l}{Combined} & \underline{3.349} & \textbf{2.597} & \multicolumn{2}{l}{$-0.752\;[-1.279,-0.270]$}\\
\midrule
OOD15 & 432 & \multicolumn{1}{l}{GRN} & 8.213 & 6.734 & \multicolumn{2}{l}{$-1.479\;[-2.053,-0.978]$}\\
OOD15 & 432 & \multicolumn{1}{l}{Rank meet} & 7.097 & 6.127 & \multicolumn{2}{l}{$-0.970\;[-1.472,-0.560]$}\\
OOD15 & 432 & \multicolumn{1}{l}{Combined} & \underline{4.523} & \textbf{3.349} & \multicolumn{2}{l}{$-1.174\;[-1.500,-0.865]$}\\
\midrule
OOD20 & 574 & \multicolumn{1}{l}{GRN} & 11.027 & 8.028 & \multicolumn{2}{l}{$-2.999\;[-4.372,-1.879]$}\\
OOD20 & 574 & \multicolumn{1}{l}{Rank meet} & 9.319 & 7.608 & \multicolumn{2}{l}{$-1.711\;[-2.608,-0.995]$}\\
OOD20 & 574 & \multicolumn{1}{l}{Combined} & \underline{6.332} & \textbf{3.873} & \multicolumn{2}{l}{$-2.459\;[-3.225,-1.774]$}\\
\midrule
\multicolumn{7}{l}{\textbf{(b) Population-based search}}\\
Split & Targets & CDGS--GRN & CDGS--static & CDGS--refresh & Combined refresh (direct ranking) & $\Delta$ CDGS $[95\%\,\mathrm{CI}]$\\
\midrule
OOD10 & 129 & 5.447 & 4.184 & \underline{3.047} & \textbf{2.597} & $-1.138$ $[-2.145,-.341]$\\
OOD15 & 432 & 8.515 & 5.844 & \underline{3.828} & \textbf{3.349} & $-2.016$ $[-2.612,-1.463]$\\
OOD20 & 574 & 8.482 & 7.489 & \underline{4.158} & \textbf{3.873} & $-3.331$ $[-4.224,-2.493]$\\
\midrule
\multicolumn{7}{l}{\textbf{(c) Scored states and sampled sequence positions per target}}\\
Split & Targets & \shortstack{Combined refresh\\GRN states} & \shortstack{CDGS--refresh\\GRN states} & \shortstack{CDGS--refresh\\scored positions} & \shortstack{CDGS--GRN\\scored positions} & \shortstack{State count\\reduction factor}\\
\midrule
OOD10 & 129 & \textbf{2.00} & \underline{12.96} & 144.85 & 353.33 & 6.48\\
OOD15 & 432 & \textbf{2.64} & \underline{23.64} & 214.67 & 624.21 & 8.96\\
OOD20 & 574 & \textbf{3.14} & \underline{38.32} & 258.60 & 669.05 & 12.22\\
\bottomrule
\end{tabular}}
\end{table}

Across all three scales, Combined, rank meet, and CDGS-style search reduce mean query scores against matched Static controls, with paired CIs below zero (Table~\ref{tab:cdgs-main}).  SymBuild-style rank meet lowers mean scores by $11.30$--$18.36\%$ and the OOD20 95th-percentile score from $25$ to $17$.  Tail scores also fall on OOD10 and OOD15 (Appendix Table~\ref{tab:grn-direct-diagnostic}).  Rank meet refresh also lowers mean query scores in CDGS-style search by $12.48$--$19.10\%$ across scales, with paired gains and scoring costs in Appendix Tables~\ref{tab:cdgs-aggregation-cross} and~\ref{tab:cdgs-aggregation-cost}.  Direct Combined refresh achieves the lowest means with aggregation suited to domain signals and requires $6.48$--$12.22$ times fewer distinct scored states than CDGS--refresh.  Including initial scoring, its query savings offset additional scoring cost above verifier latencies of $12.55$--$31.44$ ms per query (Appendix Tables~\ref{tab:timing-full} and~\ref{tab:latency-grid}).

\section{Discussion and Limitations}

Terminal correspondence reuses directed process knowledge across equivalent outcomes, while statewise refinement adapts its relevance to execution history.  Results across aggregation, planning, and learned scoring support domain-specific combinations of process and state signals, exemplified by Combined's query efficiency in GRN target removal.  The framework is especially useful when reliable correspondence and reusable process evidence guide decisions with costly verification.  Correspondence fidelity governs the quality of transported information, and verifier reliability governs execution certification (Appendix~\ref{app:robustness}).  Under Theorem~\ref{thm:complete}'s assumptions and sufficient budget, rank meet guarantees completion with a tight worst-case query bound through union coverage and suffix preservation (Appendix~\ref{app:completion-proof}).  Future work will explore aggregation adapted to domain signals and realized states, and joint reasoning over multiple terminal correspondences.
\section{Conclusion}
Terminal correspondence transports process directionality; realized state updates its relevance.  \textup{SymBuild} makes this separation analyzable and improves verified efficiency in three construction domains, with its ordinal adaptation also improving target removal.  Framework tests cover four domains, with refresh gains under alternative aggregation, planning, and learned scoring.  In short, transport what the outcome preserves, refine what history changes, and leave acceptance to the verifier.
\section*{Reproducibility Statement}
Appendices~\ref{app:extended-evidence} and~\ref{app:setup} report controls, results, protocols, and the computing environment.  The complete code will be released in due course for community evaluation.

\clearpage
\bibliography{references}
\bibliographystyle{plainnat}

\clearpage
\appendix
\section*{Appendices}
\startcontents[appendix]
\printcontents[appendix]{}{1}{\setcounter{tocdepth}{3}}

\section{Proofs}
\label{app:proofs}

\subsection{Proof of Theorem~\ref{thm:nonident}}

Let both systems use the action set $\{a,b\}$ and the same terminal object $o$ with a nonidentity automorphism $g$ that exchanges two indistinguishable terminal roles, so $g\cdot o=o$ and the terminal stabilizers coincide.  In $\mathcal C_{\parallel}$, define the diamond transitions $s_0\xrightarrow{a}s_a\xrightarrow{b}s_T$ and $s_0\xrightarrow{b}s_b\xrightarrow{a}s_T$; then $a$ and $b$ commute at $s_0$.  In $\mathcal C_{\rightarrow}$, define only $q_0\xrightarrow{a}q_a\xrightarrow{b}q_T$, so $a\prec b$ is forced.  Thus the same nontrivial terminal symmetry is compatible with different process relations.  Let the behavior policy emit $ab$ with probability one in both systems.  The observation distributions are therefore identical whether the estimator sees only the terminal object or also one demonstration.

Fix the shared observation $z$.  If a randomized estimator outputs the first relation with probability $p$ and the second with probability $q$, then $p+q\le1$.  Its errors on the two systems are at least $1-p$ and $1-q$, hence
\begin{equation}
 \max\{1-p,1-q\}\ge 1-\frac{p+q}{2}\ge\frac12.
\end{equation}
Thus no universally consistent estimator exists.  Repeated terminal observations preserve the same point-mass observation distribution.

\subsection{Proof of Lemma~\ref{lem:union}}

For every action $a$,
\begin{equation}
 m_s(a)\le k
 \iff \min\{r_T^s(a),r_X^s(a)\}\le k
 \iff r_T^s(a)\le k\ \text{or}\ r_X^s(a)\le k.
\end{equation}
This proves the set identity.  Its cardinality is at most $B_T^s(k)+B_X^s(k)$.  If a feasible action is in the union, an ordering that exhausts meet ranks through $k$ must encounter one within that many queries.  For strict ranks and integer $1\le k\le|A_s|$, $B_T=B_X=k$.

\subsection{Proof of Theorem~\ref{thm:complete}}
\label{app:completion-proof}

At each incomplete state, Lemma~\ref{lem:union} places every member of $U_s(k_s)$ before every action outside it.  Coverage supplies an accepted member within $|U_s(k_s)|$ calls.  Extension preservation leaves a complete suffix after the transition.  The horizon bound therefore gives completion in $T\le H$ accepted transitions.  Summing the union sizes over the visited states $s_0,\ldots,s_{T-1}$ proves the verifier-call bound.  A finite cap returns the current trajectory when calls are exhausted.  With $k_s=|A_s|$, $U_s(k_s)=A_s$, giving relative completeness.

For optimality, fix a state and any deterministic query policy.  An adversary returns infeasible for every action outside $U_s(k)$ and for queried union members until its last unqueried member, which is feasible.  Repeated queries retain their previous labels.  This satisfies the coverage promise and forces at least $|U_s(k)|$ calls; a policy that never exhausts the union cannot certify success.  Meet queries the union before outside actions and matches the lower bound.

\paragraph{Verifier safety and covariance.}
Acceptance is determined by the fixed verifier for every proposal order.  If the selected correspondence induces a candidate bijection $\varphi_g:A_s\to A_{gs}$ and both ranks are covariant, with $r_T^{gs}(\varphi_g(a))=r_T^s(a)$ and $r_X^{gs}(\varphi_g(a))=r_X^s(a)$, then $m_{gs}(\varphi_g(a))=m_s(a)$.  Here $\varphi_g$ includes the domain's entity and action-parameter maps.  Equality among orbit actions within one realized state additionally requires stabilizer invariance; covariance concerns transformed states.

\subsection{Random interruption identity}

Conditioning on the independently drawn budget gives
\begin{equation}
 \Pr(Q_\pi\le B)
 =\sum_{b\in\mathcal B}\Pr(B=b)\Pr(Q_\pi\le b)
 =\sum_{b\in\mathcal B}\mu(b)F_\pi(b).
\end{equation}
Pointwise cumulative distribution function (CDF) dominance implies dominance for every nonnegative weighting.  Conversely, choosing a point mass at each $b$ recovers each pointwise inequality.

\subsection{Proof of Proposition~\ref{prop:strict}}
\label{app:refresh-separation}

\begin{proposition}[Refresh versus static separation]
\label{prop:strict}
There exists a problem with two nonterminal decision states in which refreshed and static policies use the same terminal prior, fixed verifier, and first action, yet refresh completes in two calls while static is interrupted at budget two.
\end{proposition}

Let $s_0$ have candidates $p,a,b$ and accept only $p$, leading to $s_1$.  At $s_1$, $a$ is rejected and $b$ is accepted, completing the task.  Each query costs one call.  The ranks and their meet are
\begin{center}
\begin{tabular}{llrrr}
\toprule
State & Action & $r_T$ & $r_X$ & $\min(r_T,r_X)$\\
\midrule
$s_0$ & $p$ & 1 & 1 & 1\\
$s_0$ & $a$ & 2.5 & 2 & 2\\
$s_0$ & $b$ & 2.5 & 3 & 2.5\\
$s_1$ & $a$ & 1.5 & 2 & 1.5\\
$s_1$ & $b$ & 1.5 & 1 & 1\\
\bottomrule
\end{tabular}
\end{center}
The fixed process scores tie $a,b$; restricting them to remaining candidates changes their average rank from $2.5$ to $1.5$.  Static caches the initial order $p,a,b$ and retains $a,b$ after accepting $p$.  Refresh orders $b,a$ at $s_1$.  Budget two therefore gives the traces $p,a$ for Static and $p,b$ for refresh.  Both decisive orders follow from the primary meet ranks.

\subsection{Capped cost identity}

For dense integer budgets, anytime completion is also linked to capped time to solution.  For a nonnegative integer $q$ and cap $M$, $\min\{q,M\}=\sum_{b=0}^{M-1}\ind\{q>b\}$.  Taking expectations and using $\Pr(Q>b)=1-F(b)$ gives
\begin{equation}
  1-\frac{\mathbb E[\min\{Q_\pi,M\}]}{M}
  =\frac{1}{M}\sum_{b=0}^{M-1}F_\pi(b).
  \label{eq:layercake}
\end{equation}

\section{Operational Relation to Prior Symmetry Uses}
\label{app:recent-symmetry}

Table~\ref{tab:recent-symmetry} compares the symmetry locus and information role of the closest historical and modern precedents.  Classical planning and SAT expose complementary forms of reuse.  \citet{fox2005action} use almost symmetry derived from abstraction to guide forward action ordering.  In SAT, known automorphisms preserve logical consequences: symmetry propagation derives implications, and symmetric explanation learning maps explanation clauses to consequences useful under the current partial assignment \citep{devriendt2012symprop,devriendt2017sel}.  Our framework transports independently identified process evidence through terminal correspondence.  The realized state refines this proposal information, and the fixed verifier determines action acceptance.

The broader landscape includes MDP homomorphic networks and equivariant Q learning methods that encode joint state-action symmetry in value and policy structure \citep{vanderpol2020mdp,wang2022eqq}, soft or approximate equivariance under misspecified symmetries \citep{finzi2021rpp,wang2022approx}, and graph generation methods that incorporate symmetry \citep{zhao2024pard,kim2025sagfn}.  Across these settings, the carrier framework separates what is supplied by process evidence, what terminal correspondence can carry across equivalent outcomes, and what must be updated from the realized state.

\begin{table}[h]
\centering
\caption{\textbf{Operational comparison of symmetry uses.}  The key distinction is where symmetry is assumed and what information it contributes to decision making.}
\label{tab:recent-symmetry}
\scriptsize
\renewcommand{\arraystretch}{0.90}
\setlength{\tabcolsep}{2.4pt}
\begin{tabular}{p{.16\linewidth}p{.18\linewidth}p{.18\linewidth}p{.20\linewidth}p{.20\linewidth}}
\toprule
Work & Symmetry locus & Primary use & Process assumption or information source & Operational relation\\
\midrule
\citet{liu1990assembly} & Assembly component geometry & Kinematic and spatial constraint inference & Component feature symmetries reduce constraint satisfaction problem (CSP) combinatorics & The assembly method uses component symmetry for constraint inference; our framework transports sequential process relations\\
\citet{fox1999symmetry} & Planning problem objects and actions & Detect and exploit symmetric search structure & Exact problem symmetry exposes equivalent choices & The planner reduces equivalent choices; our framework transports process directionality obtained from independent evidence\\
\citet{fox2005action} & Abstract planning problem; objects and actions that are almost symmetric & Proactive action ordering in forward search & Guidance is induced from symmetry revealed by abstraction and prior plan actions & The planner uses abstract symmetry for action ordering; our framework transports independently identified process relations\\
Dynamic SAT symmetry handling \citep{devriendt2012symprop,devriendt2017sel} & Boolean theory and literal symmetries & State-dependent propagation and clause learning & Full problem symmetry preserves logical consequences; the current assignment determines when transported consequences restrict search & SAT transports logical consequences under formula symmetry; our framework transports process evidence and refines proposal ranks\\
SayCan \citeyearpar{ichter2023saycan} & Embodied state and skill set & Ground language model skill selection in current affordances & Value functions score whether each pretrained skill is currently executable and are reevaluated through execution & SayCan reevaluates skill affordances after state changes; our framework adds terminal transport of independently identified process structure\\
SymPlan \citeyearpar{zhao2023symplan} & Path planning Markov decision process (MDP) and value iteration operator & Equivariant differentiable planning & Relevant state and action transformations are symmetries of the planning problem & SymPlan structures planner computation by equivariance; our framework separates process evidence, terminal correspondence, and state refinement\\
Equivariant Motion Manifold Primitives \citeyearpar{lee2023emmp} & Manifold of tasks and solution trajectories & Equivariant trajectory family learning & Task symmetry supports equivariant motion families & The motion model learns equivariant trajectory families; our framework combines terminal correspondence, process evidence, and state residuals\\
Partial equivariant RL \citeyearpar{chang2026partially} & State-action regions of an MDP & Selective invariant and standard Bellman backups & Symmetry validity varies across local process regions & Partial equivariant RL adapts Bellman backups locally; our framework uses fixed terminal correspondence and updated decision relevance\\
Cross-space symmetry composition \citeyearpar{hadjiloizou2026crossspace} & Configuration and task spaces & Lifting, descending, and composing symmetries for jointly equivariant policies & Multiple policy symmetries are composed across spaces & The policy composes configuration and task symmetries; our framework transports directed process evidence through terminal correspondence\\
Canonical diffusion \citeyearpar{zhou2026canonical} & Orbit representatives of invariant targets & Canonical representative selection & Invariant target distribution is modeled on a canonical slice & Canonical diffusion selects an orbit representative; our framework uses state residuals to update the transported structure's decision value\\
Carrier framework & Terminal outcomes plus external process evidence & Transport, refinement, and certification for verified construction & Directionality is independently grounded; exact terminal correspondence supplies transport & Our framework fixes transported structure within an episode, updates its decision relevance, and certifies actions with a fixed verifier\\
\bottomrule
\end{tabular}
\end{table}

\FloatBarrier

\section{Extended Empirical Evidence}
\label{app:extended-evidence}

\subsection{\textup{SymBuild} Baselines and Ablations in Three Construction Domains}
\label{app:complete-tables}

\begin{algorithm}[h]
\caption{\textup{SymBuild}: statewise residual rank meet}
\label{alg:symbuild}
\begin{algorithmic}[1]
\Require target $X$, correspondence $g$, evidence $R_{\rm ref}$, state $s_0$, verifier $V$, call cap $B$, deterministic key $\kappa_s$
\State $R_T\gets\Call{Transport}{R_{\rm ref},g}$; $s\gets s_0$; $q\gets0$; $\tau\gets()$
\While{$s$ is incomplete and $q<B$}
  \State $A_s\gets$ remaining actions; $r_T^s\gets\Call{RestrictRank}{R_T,A_s}$
  \State $r_X^s\gets\Call{ResidualRank}{s,A_s}$; $m_s(a)\gets\min\{r_T^s(a),r_X^s(a)\}$
  \State $Q\gets\Call{Sort}{A_s,(m_s(a),\kappa_s(a))}$
  \State $\mathrm{accepted}\gets\mathrm{false}$
  \While{$Q\neq\emptyset$ and $q<B$}
    \State $a\gets\Call{PopFirst}{Q}$; $y\gets V_s(a)$; $q\gets q+1$
    \If{$y=1$}
      \State $s\gets F(s,a)$; $\tau\gets\tau\circ a$; $\mathrm{accepted}\gets\mathrm{true}$; \textbf{break}
    \EndIf
  \EndWhile
  \If{$\neg\mathrm{accepted}$ and $q<B$}
    \State \Return $(\tau,q,\text{candidates exhausted})$
  \EndIf
\EndWhile
\If{$s$ is complete}
  \State \Return $(\tau,q,\text{completed})$
\Else
  \State \Return $(\tau,q,\text{budget exhausted})$
\EndIf
\end{algorithmic}
\end{algorithm}

\begin{table}[h]
\centering
\caption{\textbf{Formal \textup{SymBuild} evaluation coverage in the three construction domains.}  Counts refer to confirmation objects; each seed is evaluated for every applicable method and budget.  $n$ denotes parts, events, or remaining items.}
\label{tab:protocol}
\small
\setlength{\tabcolsep}{4.2pt}
\begin{tabular}{lrrrrl}
\toprule
Domain & Objects & Seeds & Setting & Budget grid & Verifier resource\\
\midrule
CAD calls & 72 & 2 sensor & ID, 5--20 parts & 6 values, $1.25n$--$4n$ & proposal calls\\
CAD work & 48 & 2 sensor & ID, 5--20 parts & 6 values, $1.25$--$4$ & dense sweep steps\\
Mini-Program & 480 & 2 data & ID, length OOD & 5 values, $1n$--$2n$ & precedence queries\\
Exact-fill packing & 480 & 2 data & ID, size OOD & 6 values, $1n$--$3n$ & exact extension queries\\
\bottomrule
\end{tabular}
\end{table}

Table~\ref{tab:complete-matrix} compares \textup{SymBuild}, its single-channel ablations, and permuted transport.  Alternative aggregation and planning methods appear in Appendices~\ref{app:aggregation-transfer} and~\ref{app:search-controls}.  The process and state controls are, respectively, PCS-ICP and outsideness for CAD, combined learned and observed process relations and transported observed relations for Mini-Programs, and the process prior and best fit for exact-fill packing.  CAD correspondence evidence comes from the aligned versus permuted intervention on disjoint objects in Table~\ref{tab:ablation}(b); the corresponding cells in Table~\ref{tab:complete-matrix} are marked with an em dash.

\begin{table}[H]
\centering
\caption{\textbf{Complete \textup{SymBuild} comparison and ablation matrix.}  Each cell gives anytime AUC first and final verified success second, both in percent ($\uparrow$).  Static is the matched static control for statewise refresh.  Process and State retain only the corresponding information channel; Permuted disrupts transport while preserving the remaining interface.  Program OOD and Packing OOD denote the length and size OOD splits.  Best and second-best distinct AUCs among the displayed methods are bold and underlined; ties share formatting.}
\label{tab:complete-matrix}
\scriptsize
\setlength{\tabcolsep}{3.0pt}
\begin{tabular}{llrrrrr}
\toprule
Panel & $N$ & \textup{SymBuild} & Static & Process & State & Permuted\\
\midrule
CAD calls & 72 & \textbf{81.60/94.62} & 78.73/89.93 & \underline{79.57/89.58} & 73.12/91.49 & --\\
CAD work & 48 & \textbf{98.70/98.70} & \underline{91.93/91.93} & 87.50/87.50 & 85.63/85.94 & --\\
Program ID & 240 & \underline{85.08/100} & 72.67/100 & 84.83/100 & \textbf{85.17/100} & 15.83/37.50\\
Program OOD & 240 & \underline{64.67/97.50} & 42.92/83.33 & 63.58/97.08 & \textbf{65.83/97.50} & 1.25/4.58\\
Packing ID & 240 & \textbf{57.92/100} & 49.38/99.58 & \underline{50.21/100} & 38.26/100 & 13.33/68.75\\
Packing OOD & 240 & \textbf{34.03/95.00} & 25.35/83.33 & \underline{28.40/93.75} & 21.67/99.17 & 3.68/20.42\\
\bottomrule
\end{tabular}
\end{table}

Table~\ref{tab:cad-mechanism-ci} reports paired uncertainty for the CAD mechanism contrasts in Main Table~\ref{tab:ablation}(b); Appendix~\ref{app:setup} specifies the panel construction and resampling rules.

\begin{table}[h]
\centering
\caption{\textbf{Paired uncertainty for Main Table~\ref{tab:ablation}(b).}  Success gain is the named condition minus its comparator; call saving is the comparator minus the named condition in calls per part, so positive values favor the named condition.  Intervals are paired 95\% CIs from 10,000 bootstrap draws over assemblies; Appendix~\ref{app:setup} specifies the panel construction and resampling units.  All reported confidence intervals for call savings are positive.}
\label{tab:cad-mechanism-ci}
\scriptsize
\setlength{\tabcolsep}{4.0pt}
\begin{tabular}{lllrr}
\toprule
Test & Condition & Comparator & Success gain (pp $\uparrow$) [95\% CI] & Call saving $\uparrow$ [95\% CI]\\
\midrule
Algorithm & \textup{SymBuild} & PCS-ICP & $+5.00\,[-.42,10.83]$ & $+.143\,[.050,.240]$\\
Algorithm & \textup{SymBuild} & outsideness & $+7.71\,[2.29,13.54]$ & $+.168\,[.066,.279]$\\
Timing & \textup{SymBuild} & Static & $+4.38\,[-.42,9.58]$ & $+.152\,[.064,.246]$\\
Alignment & aligned & permuted & $+16.25\,[8.33,25.00]$ & $+.637\,[.501,.782]$\\
\bottomrule
\end{tabular}
\end{table}

\FloatBarrier
\subsection{\textup{SymBuild} Replication by Group}
\label{app:mechanism-replication}

\begin{table}[H]
\centering
\caption{\textbf{AUC replication for \textup{SymBuild} relative to Static.}  Values report the AUC difference, computed as \textup{SymBuild} minus Static; brackets are paired $95\%$ intervals.  Within each row, paired differences and intervals follow the order of the two groups from left to right.}
\label{tab:replication}
\scriptsize
\renewcommand{\arraystretch}{0.76}
\setlength{\tabcolsep}{3.6pt}
\begin{tabular}{llrr}
\toprule
Domain and resource & Group & Gap (pp $\uparrow$) & $95\%$ CI\\
\midrule
CAD calls & sensor 701 / 709 & $+2.55 / +3.18$ & $[-.29,5.44] / [.12,6.31]$\\
CAD calls & 5--8 / 9--20 parts & $+.61 / +3.99$ & $[-5.12,6.34] / [1.00,7.29]$\\
CAD work & sensor 811 / 823 & $+6.77 / +6.77$ & $[1.04,14.06] / [1.04,14.06]$\\
CAD work & 5--8 / 9--20 parts & $+3.91 / +8.20$ & $[0,11.72] / [.39,17.97]$\\
Program ID & seed 20270117 / 20270129 & $+10.50 / +14.33$ & $[6.50,14.50] / [10.17,18.67]$\\
Program OOD & seed 20270117 / 20270129 & $+24.50 / +19.00$ & $[19.50,29.33] / [14.67,23.33]$\\
Packing ID & seed 20271103 / 20271117 & $+8.06 / +9.03$ & $[5.83,10.42] / [6.81,11.25]$\\
Packing OOD & seed 20271103 / 20271117 & $+7.36 / +10.00$ & $[5.28,9.58] / [7.64,12.64]$\\
\bottomrule
\end{tabular}
\end{table}

\FloatBarrier
\subsection{Correspondence Fidelity and Approximate Verification}
\label{app:robustness}
This section expands Section~\ref{sec:boundary-main} with the graded correspondence fidelity intervention and the complete approximate verifier sweep.

\paragraph{Mini-Program intervention on correspondence fidelity.}
To diagnose correspondence fidelity, we progressively mismatch action identities while reusing the same frozen 480-object confirmation population, checkpoint, observed expert, verifier, and budget grid.  For a requested fraction $f\in\{0,.1,.25,.5,1\}$, let $M$ be the non-root event identities and set $m=\operatorname{round}(f|M|)$, except that a singleton move is promoted to two identities when possible.  A sample of $m$ identities fixed by the seed is cyclically shifted by one position and used to conjugate the learned precedence matrix; all unsampled identities remain fixed.  Thus the realized moved fraction is measured directly as $m/|M|$.  The perturbation seed is a deterministic function of data seed, split, object index, and requested mismatch level.  Intervals for the difference between clean and mismatched AUC use 10,000 paired bootstrap draws on the shared objects.

\begin{figure}[h]
  \centering
  \includegraphics[width=.82\textwidth]{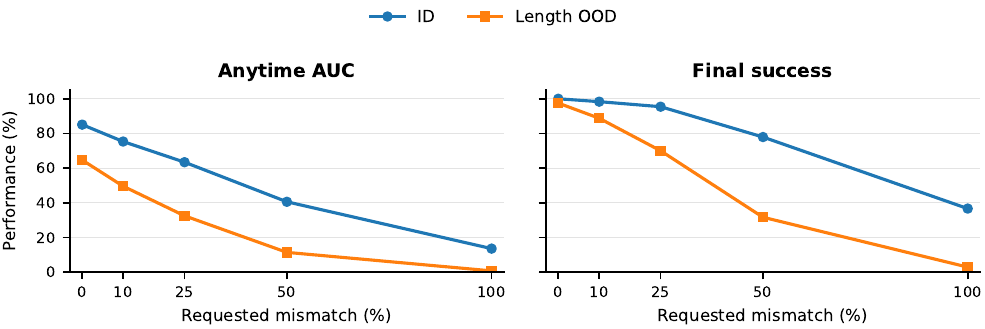}
  \caption{\textbf{Transport value decreases as correspondence fidelity degrades.}  Exact AUC and final success values appear in Main Table~\ref{tab:boundary-main}(a); both decrease as transported action identities are increasingly mismatched.}
  \label{fig:correspondence-noise}
\end{figure}

At the requested $10\%$ mismatch, moved fractions are $12.78\%$ on ID and $10.21\%$ on OOD because small graphs require a swap of two nodes; all other means are within $.74$ points of the request.  The difference in AUC between clean and mismatched conditions is already positive at the first nonzero level: $9.75$ $[7.75,11.83]$ points on ID and $15.08$ $[12.67,17.50]$ on length OOD.  The monotone frontier in Figure~\ref{fig:correspondence-noise} shows that transport value changes systematically with correspondence fidelity.

\FloatBarrier
\input{approx_verification_appendix}

\FloatBarrier
\subsection{Complete Anytime Frontiers}
\label{app:frontiers}

Table~\ref{tab:all-frontiers} gives the exact success percentages for rank meet with statewise refresh and its Static control underlying Figure~\ref{fig:frontiers}.  The ablations using a single channel appear in Main Table~\ref{tab:ablation}(a), while the independent CAD alignment intervention is summarized in Main Table~\ref{tab:ablation}(b).

\begin{table}[h]
\centering
\caption{All budget points for rank meet with statewise refresh and its Static control.  Budgets were predeclared and normalized by the number of parts, events, or remaining items, or by one dense sweep per part.  Best values are bold and second-best values are underlined; ties are unmarked.}
\label{tab:all-frontiers}
\scriptsize
\setlength{\tabcolsep}{3.0pt}
\begin{tabular}{lll}
\toprule
Domain and split & Predeclared budgets & Verified success $\uparrow$ at successive budgets\\
\midrule
CAD calls & $1.25,\,1.5,\,2,\,2.5,\,3,\,4$ & \textup{SymBuild}: $\mathbf{68.92},\,\mathbf{74.13},\,78.65,\,\mathbf{83.16},\,\mathbf{90.10},\,\mathbf{94.62}$\\
& & Static: $\underline{64.93},\,\underline{71.35},\,78.65,\,\underline{82.47},\,\underline{85.07},\,\underline{89.93}$\\
CAD work & $1.25,\,1.5,\,2,\,2.5,\,3,\,4$ & \textup{SymBuild}: $\mathbf{98.70},\,\mathbf{98.70},\,\mathbf{98.70},\,\mathbf{98.70},\,\mathbf{98.70},\,\mathbf{98.70}$\\
& & Static: $\underline{91.93},\,\underline{91.93},\,\underline{91.93},\,\underline{91.93},\,\underline{91.93},\,\underline{91.93}$\\
Program ID & $1,\,1.25,\,1.5,\,1.75,\,2$ & \textup{SymBuild}: $\mathbf{48.33},\,\mathbf{82.92},\,\mathbf{94.58},\,\mathbf{99.58},\,100.00$\\
& & Static: $\underline{27.92},\,\underline{55.42},\,\underline{82.92},\,\underline{97.08},\,100.00$\\
Program OOD & $1,\,1.25,\,1.5,\,1.75,\,2$ & \textup{SymBuild}: $\mathbf{8.33},\,\mathbf{47.92},\,\mathbf{78.75},\,\mathbf{90.83},\,\mathbf{97.50}$\\
& & Static: $\underline{5.42},\,\underline{18.33},\,\underline{40.83},\,\underline{66.67},\,\underline{83.33}$\\
Packing ID & $1,\,1.125,\,1.25,\,1.5,\,2,\,3$ & \textup{SymBuild}: $9.58,\,\mathbf{31.25},\,\mathbf{44.17},\,\mathbf{72.92},\,\mathbf{89.58},\,\mathbf{100.00}$\\
& & Static: $9.58,\,\underline{22.50},\,\underline{32.92},\,\underline{50.42},\,\underline{81.25},\,\underline{99.58}$\\
Packing OOD & $1,\,1.125,\,1.25,\,1.5,\,2,\,3$ & \textup{SymBuild}: $0.00,\,\mathbf{4.58},\,\mathbf{16.25},\,\mathbf{30.42},\,\mathbf{57.92},\,\mathbf{95.00}$\\
& & Static: $0.00,\,\underline{2.50},\,\underline{5.42},\,\underline{16.67},\,\underline{44.17},\,\underline{83.33}$\\
\bottomrule
\end{tabular}
\end{table}

\FloatBarrier
\input{cad_work_fine_appendix}
\FloatBarrier
\subsection{Alternative Aggregation Beyond \textup{SymBuild}}
\label{app:aggregation-transfer}
\input{aggregation_transfer_appendix}

\begin{table}[h]
\centering
\caption{\textbf{Post-transition pattern with alternative aggregators.}  Every pair agrees on its first accepted action.  Sequence disagreement comes from matched execution traces; gaps compare refresh with the identical static aggregation.  Product denotes the SayCanPay-adapted product rule; the Refresh only and Static only columns count verified completions unique to the refreshed and static variants, respectively.}
\label{tab:refresh-table}
\scriptsize
\renewcommand{\arraystretch}{0.76}
\setlength{\tabcolsep}{3.2pt}
\resizebox{\textwidth}{!}{%
\begin{tabular}{llrrrrl}
\toprule
Aggregator & Panel & Pairs & First same & Sequence differs & Refresh only / Static only & AUC gap $\uparrow$ $[95\%\,\mathrm{CI}]$\\
\midrule
Rank meet & CAD calls & 576 & 100\% & 96.35\% & 32 / 5 & $+2.86$ $[.12,5.76]$\\
Rank meet & CAD work & 384 & 100\% & 93.75\% & 26 / 0 & $+6.77$ $[1.30,14.06]$\\
Rank meet & Packing ID & 240 & 100\% & 67.50\% & 1 / 0 & $+8.54$ $[7.01,10.14]$\\
Rank meet & Packing OOD & 240 & 100\% & 93.75\% & 29 / 1 & $+8.68$ $[7.01,10.42]$\\
Product & CAD calls & 576 & 100\% & 97.40\% & 36 / 5 & $+3.30$ $[1.01,5.79]$\\
Product & CAD work & 384 & 100\% & 94.79\% & 17 / 0 & $+4.43$ $[.52,9.90]$\\
Product & Program ID & 240 & 100\% & 98.75\% & 0 / 0 & $+12.42$ $[9.50,15.33]$\\
Product & Program OOD & 240 & 100\% & 100\% & 36 / 3 & $+22.58$ $[19.33,25.83]$\\
Product & Packing ID & 240 & 100\% & 97.08\% & 0 / 0 & $+6.46$ $[4.51,8.47]$\\
Product & Packing OOD & 240 & 100\% & 100\% & 0 / 2 & $+6.53$ $[4.58,8.47]$\\
\bottomrule
\end{tabular}}
\end{table}

\paragraph{Post-transition pattern with alternative aggregators.}
The product rule provides a second aggregation test of the same refresh mechanism in the three construction domains.  Table~\ref{tab:refresh-table} reports the paired post-transition pattern for rank meet and product variants under matched static controls.

\FloatBarrier
\input{search_controls_appendix}

\FloatBarrier
\subsection{GRN and Population-Based Search}
\label{app:grn-search}

\paragraph{Target episodes.}
The official Panda OOD scenes \citep{bouhsain2025grn} are converted into target removal episodes with one deterministic annotation verifier shared by all methods.  From the empty removed set, breadth-first search enumerates states reachable through accepted removals; a movable object is a target if it is removed on at least one reachable transition.  This yields 129 targets from 20 OOD10 scenes, 432 from 50 OOD15 scenes, and 574 from 50 OOD20 scenes, for 1,135 targets total.

\paragraph{Candidate verification and transitions.}
Every remaining movable object contributes five candidates in the fixed grasp order Top, Front, Rear, Right, Left.  Candidate $(o,g)$ is accepted iff the object remains annotated, no remaining object names $o$ as its parent in the support frame, the official inverse kinematics (IK) annotation is feasible, and the sum of geometric obstruction failure ratios from remaining blockers is below $1-10^{-12}$.  Acceptance removes the object and clears rejected candidates; rejection leaves the scene unchanged and suppresses only that object and grasp pair until the next accepted removal.

\paragraph{Score construction.}
The fixed GRN model assigns a grasp feasibility score to each candidate object and grasp pair.  Process utility comes from the current directed obstruction and support graph: reverse transitive depth from the target gives larger values to remaining blocker ancestors and $-1$ to objects outside the target ancestry, with each object value copied to its five grasps.  Each channel is independently normalized to z scores over the current $(\text{object},\text{grasp})$ pairs; a channel with standard deviation below $10^{-12}$ is mapped to zero.  Ranking by GRN or process uses the corresponding normalized channel, and combined ranking uses their unweighted sum.

\paragraph{Static and refresh updates.}
GRN static caches the initial raw GRN values. Combined static caches both initial raw channels.  After accepted removals, removed objects are filtered out and the cached channels are normalized again over the remaining candidates.  GRN refresh recomputes the raw GRN values after each accepted removal. Combined refresh recomputes both raw channels before normalization.  Candidates are ordered by descending score with a shared deterministic tie order.

\paragraph{Budgets and statistical evaluation.}
All methods use a $5N$ verifier call limit. Table~\ref{tab:cdgs-main} records issued calls on success and $5N+1$ on failure.  Direct ranking uses tie seeds 0--4 at all three scales.  Results are averaged over seeds for each target before inference, which uses 20,000 bootstrap draws clustered by scene and sign permutations with seed 20260809.

\paragraph{Population-based search.}
Our adaptation follows the population-based search strategy of \citet{mishra2026cdgs} under the same candidate, score, transition, and verifier interface.  Each of two rounds samples 12 imagined sequences of horizon at most four from a softmax with temperature $.8$; imagined removals use no verifier calls.  A sequence receives its log sampling probability divided by sequence length plus a bonus of 2 if it reaches the target, and the top $25\%$ form the elite set.  Between rounds, counts of first actions in the elite set are weighted by inverse rank to form a guide distribution; its log probability enters the score at the first step with weight $.5$.  After the second round, the first action with the largest exponentiated elite vote is passed to the shared verifier.  CDGS--GRN uses the refreshed GRN channel alone. CDGS--static uses the static combined channels above. CDGS--refresh uses the refreshed combined channels.  Population-based search uses seeds 0--4 at all three scales.  The seed sets are shared across variants; samples follow each variant's score distribution.  Target averaging and inference match the direct methods, including clustering by scene.

\input{grn_ordinal_appendix}

\FloatBarrier
\input{grn_cdgs_ordinal_appendix}

\paragraph{Computational cost of population-based search.}
Distinct scored removal states, including the initial state, measure the demand on the neural scorer; action positions evaluated in imagined sequences measure the additional search effort.  During effectiveness evaluation, predictions for identical scenes and removed sets are reused across methods, targets, and seeds.  Each method's scoring requirements are counted independently of this reuse.  CDGS--GRN evaluates $43.73$, $71.26$, and $109.06$ distinct GRN states per target on OOD10, OOD15, and OOD20, respectively.  Combined refresh and CDGS--refresh both achieve $100\%$ completion under the shared $5N$ cap.  Their state counts in Table~\ref{tab:cdgs-main}(c) show that direct ranking attains this coverage with substantially less neural scoring than population-based search.

\begin{table}[h]
\centering
\caption{\textbf{Related methods and their roles in this paper.}  GRN and SayCanPay-adapted numerical comparisons use matched candidates, verifier, and budgets. Planning Neural Operator (PNO) supplies a motion-planning reference.}
\label{tab:open-code-compare}
\scriptsize
\setlength{\tabcolsep}{3.0pt}
\begin{tabular}{p{.17\textwidth}p{.10\textwidth}p{.15\textwidth}p{.25\textwidth}p{.25\textwidth}}
\toprule
Method & Venue & License & Native interface & Use in this paper\\
\midrule
SayCanPay \citep{hazra2024saycanpay} & AAAI 2024 & not stated in the official repository & language actions with Can and Pay scores & matched refresh and static comparison using the product rule\\
GRN \citep{bouhsain2025grn} & ICLR 2025 & MIT & 3D task and motion planning graph; grasp, inverse kinematics, and geometric obstruction labels & official OOD scorer and comparison under matched search\\
PNO \citep{matada2025pno} & ICLR 2025 & CC BY-NC-SA 4.0 & continuous value field over maps and configurations & contextual comparison at the task and interface levels\\
\bottomrule
\end{tabular}
\end{table}

\FloatBarrier
\input{grn_timing_appendix}

\FloatBarrier

\section{Experimental Protocols and Environment}
\label{app:setup}

\paragraph{Data provenance and domain construction.}
CAD objects come from the official Automated Sequence Planning (ASAP) training partition \citep{tian2024asap}; each confirmation panel is disjoint from earlier CAD experiments in both object identity and mesh content before deterministic selection.  Mini-Programs are generated deterministically from the data seed and latent program specification, with confirmation splits that are disjoint in identity.  Exact-fill packing instances are generated deterministically from the data seed, split, weights, true groups, the transport mapping from groups to bins, and visible anchors; confirmation IDs are pairwise disjoint and separate from development.  Generation and ranking rules for each domain follow below.

\subsection{Shared Evaluation and Statistics}
\label{app:shared-evaluation}
\paragraph{Paired comparisons.}
The six primary CAD, Mini-Program, and packing panels use $10{,}000$ bootstrap draws paired at the object or assembly level; GRN and CDGS-style intervals use $20{,}000$ resamples clustered by scene after averaging search seeds for each target within its scene.  Reported differences and projections are computed from unrounded values.
The construction-domain interruption grids are summarized in Table~\ref{tab:protocol}; Appendix~\ref{app:grn-search} specifies the GRN cap.

\paragraph{Rank scales and budget evaluation.}
The theory uses ranks beginning at one.  Refresh in CAD and packing applies the same positive affine normalization to both current ordinal ranks, preserving meet order and the rank threshold identity.  Algorithm~\ref{alg:symbuild} counts proposal calls.  Work-budget success requires a valid completion within the cumulative sweep threshold, evaluated from trajectories obtained under the original call cap.

\subsection{CAD Assembly}
\label{app:cad-protocol}
The CAD evaluation panels in Table~\ref{tab:cad-panel-settings} use the dense 96-step MuJoCo \citep{todorov2012mujoco} verifier based on a straight convex-hull sweep, with escape factor 1.5.  Table~\ref{tab:cad-panel-settings} collects their populations, sensor seeds, proposal caps, and bootstrap seeds.  Within an assembly, sensor and lift conditions are averaged before paired resampling.  The Algorithm and Timing contrasts share their 120 assemblies.  The other evaluation panels use disjoint assemblies as specified below.

\begin{table}[h]
\centering
\caption{\textbf{CAD evaluation panels.} Each evaluation assembly is tested under all four $C_4$ lifts. Population entries specify the parts quotas where defined. Each panel uses 10,000 bootstrap draws paired by assembly after averaging its sensor and lift conditions. Alignment training and validation populations are described in the learned residual protocol.}
\label{tab:cad-panel-settings}
\footnotesize
\setlength{\tabcolsep}{3pt}
\begin{tabular}{p{.13\textwidth}p{.36\textwidth}p{.13\textwidth}p{.15\textwidth}p{.11\textwidth}}
\toprule
Panel & Evaluation population & Sensor seeds & Proposal cap & \shortstack{Bootstrap\\seed}\\
\midrule
Calls & 72 assemblies: 24 with 5--8 parts and 48 with 9--20 parts & 701, 709 & $\lceil4N\rceil$ & 22101\\
Algorithm & 120 assemblies: 24 with 5--8 parts, 48 with 9--20 parts, and 48 with 21--50 parts & 389 & $\max(3N,24)$ & 14801\\
Timing & Same 120 assemblies as Algorithm & 503 & $\max(3N,24)$ & 14901\\
Alignment & 60 evaluation assemblies & 613 & $\max(3N,24)$ & 21501\\
Work & 48 assemblies: 16 with 5--8 parts and 32 with 9--20 parts & 811, 823 & $4N$ & 23101\\
\bottomrule
\end{tabular}
\end{table}

\paragraph{Correspondence and geometric controls.}
The external CAD correspondence controls use the official ASAP assembly assets.  Assemblies are parsed either from archive meshes already expressed in assembled terminal coordinates or from public repository examples with explicit final poses.  The development correspondence set includes finite, parseable assemblies with 5--50 parts.  A deterministic partition based on assembly identity assigns 80\% to training and 20\% to validation, and augmented views inherit the assembly split.  Terminal equivalence is the tabletop $C_4$ orbit formed by yaw rotations of $0^\circ$, $90^\circ$, $180^\circ$, and $270^\circ$ about the fixed world vertical axis.  Each correspondence view independently permutes part identities, applies a random $SE(3)$ transform in the sensor frame and one of these $C_4$ lifts, and renders directional partial observations with $55\%$ retained visibility, at most 256 points per part, and Gaussian coordinate noise $0.002$.  The gravity and tool directions expressed in the sensor frame provide the process coordinate system used by the geometric controls.

\textbf{PCS-ICP} denotes iterative closest point (ICP) correspondence in this process coordinate system with Hungarian assignment.  It first expresses the partial observations in the frame defined by the tool and gravity directions, evaluates all four $C_4$ representatives, and runs up to four translation-only trimmed ICP refinements per representative, retaining the closest $80\%$ of point correspondences at each refinement.  Trimmed one-way Chamfer costs from partial to full geometry define the matrix between observed and canonical parts; Hungarian assignment gives a one-to-one part correspondence, and the representative with minimum mean assignment cost is selected.  The recovered lift and assignment transport the canonical CAD geometry into the target frame.  At each state defined by the remaining parts and for each fixed world direction, a directional blocker relation is recomputed on the predicted parts.  The transported process score for a candidate is its number of predicted blockers plus a $-0.01$ tie-break based on normalized outward projection, with lower scores queried first.  The \textbf{outsideness} control adapts ASAP's outside-in part heuristic \citep{tian2024asap} to the shared action set that pairs each part with a removal direction: using the same centers aligned with the process frame, candidate actions, and verifier, it scores part $i$ along removal direction $d$ by $-c_i^\top d$.  Both controls share verifier acceptance and action semantics, isolating the proposal information.

\paragraph{Main CAD anytime confirmation on 72 objects.}
The primary CAD panel uses 72 ASAP training assemblies that are disjoint in both identity and content, with 5--20 parseable parts and a recoverable assembled configuration.  A deterministic ordering based on assembly identity selects the parts quotas in Table~\ref{tab:cad-panel-settings}.  Its two sensor seeds define base partial views; each assembly paired with each sensor seed is evaluated under all four forced $C_4$ lifts, yielding $72\times2\times4=576$ conditions per method.  Views follow the partial-observation settings in the correspondence protocol above.  One shared calibration rotation perturbs the process axes defined by the tool and gravity directions by a Gaussian angle with mean zero and standard deviation $5^\circ$.  Actions pair each remaining part with one of the six world directions ${\pm x,\pm y,\pm z}$.

For a set $R$ of remaining parts, PCS-ICP supplies the recovered lift and the assignment from observed to canonical parts used to transport the reference CAD.  For action $(i,d)$, the transported process score is the number of predicted directional blockers plus the outward tie term
\[
 s_T(i,d;R)=|\mathrm{blockers}(i,d;R)|-0.01\,\frac{\langle c_i,d\rangle}{\max\{\operatorname{range}_{j\in R}\langle c_j,d\rangle,10^{-8}\}},
\]
where $c_i$ is the transported predicted center.  The local geometric score is $s_X(i,d;R)=-\langle x_i,d\rangle$, where $x_i$ is obtained by centering the observed part centers, expressing them in the process frame aligned with the tool and gravity, and dividing all centers by their maximum radius (with a $10^{-8}$ floor).  With $m=6|R|$ current actions, each score channel is independently converted to average weak ranks and normalized to $[0,1]$ by subtracting one and dividing by $\max(m-1,1)$.  \textup{SymBuild}'s rank meet orders actions by $\min(r_T,r_X)+10^{-6}r_T+10^{-9}r_X$, followed by part and direction indices as the final deterministic tie-break.  After an accepted removal, the blocker relation and both current ranks are recomputed on the reduced action set; a rejected proposal leaves the state unchanged and is skipped for that state.  The Static control computes the complete fused ordering once at the initial state and subsequently only removes actions whose parts have already been accepted.  All methods use the common CAD verifier with penetration tolerance $\max(10^{-5}D,10^{-7})$ for assembly diameter $D$ and the call cap in Table~\ref{tab:cad-panel-settings}.  The predeclared call budgets are $\{1.25,1.5,2,2.5,3,4\}N$; AUC is their unweighted success mean; capped normalized call cost is the number of calls divided by $4N$ on success and 1 on failure; intervals follow the assembly-level protocol in Table~\ref{tab:cad-panel-settings}.

CAD relation indices are $(j,i,d)$: $B_T[j,i,d]=1$ when predicted part $j$ blocks part $i$ along direction $d$, giving a tensor with two part axes and one direction axis.  The selected assignment indexes reference geometry, and the recovered yaw lift rotates it before blocker evaluation along the six world axes.  Candidate $(i,d)$ uses its blocker column and outward projection.  The transported geometry stays fixed while relations among remaining parts and current action ranks are updated.  The frozen verifier uses the true part assignment and maps the world direction back through the true yaw lift.  At the evaluated maximum of 300 candidates, distinct normalized average ranks differ by at least $1/[2(300-1)]$, so the $10^{-6}$ and $10^{-9}$ terms only resolve meet ties.

\paragraph{Independent 120-object CAD mechanism panels.}
The Algorithm and Timing contrasts in Table~\ref{tab:ablation}(b) use the same 120 ASAP training assemblies, disjoint from earlier CAD evaluation panels in both assembly identity and mesh content.  A deterministic pseudorandom ordering selects the parts quotas in Table~\ref{tab:cad-panel-settings}.  Algorithm compares the physics oracle, PCS-ICP transport, outsideness, and \textup{SymBuild}.  Timing compares \textup{SymBuild} with Static using its independently seeded views.  Both use the common CAD verifier and the caps and bootstrap seeds in Table~\ref{tab:cad-panel-settings}.  The paired intervals in Table~\ref{tab:cad-mechanism-ci} average the four lifts within each assembly before resampling.  Alignment uses the same resampling unit over the 60 evaluation assemblies from the learned residual protocol below.  Success gain is the named condition minus its comparator, and call saving is the comparator minus the named condition.

\paragraph{Verification and complete suffixes.}
The CAD verifier returns both a Boolean decision and the number of sampled sweep positions, providing the call and work measurements.  In this fixed-pose model, remaining parts are reset to assembled poses and removed parts are moved away.  Deleting obstacles preserves a previously feasible prescribed sweep, so skipping deleted parts in a valid removal sequence gives a complete suffix.

\paragraph{CAD protocol for the learned residual.}
The aligned versus permuted CAD intervention uses mutually disjoint sets of 60 training, 20 validation, and 60 evaluation assemblies from the same public ASAP partition.  Training, validation, and evaluation use sensor seeds 601, 607, and 613, respectively, with noise of $5^\circ$ applied to the process axis.  Training and validation assign one deterministic $C_4$ lift to each assembly, while evaluation uses all four lifts.  The dense 96-step verifier labels a union of candidate probes at states visited by analytic rank meet, yielding 5,022 training and 1,460 validation state-action rows.  At each visited state, the probe union contains the top two candidates under the transported relation, outsideness, and rank meet, together with up to two candidates sampled without replacement from the remainder.  A multilayer perceptron (MLP) with two hidden layers of width 64 is trained for 2,000 AdamW steps with seed 307, learning rate $10^{-3}$, batch size 256, and validation every 100 steps; validation binary cross-entropy (BCE) selects the checkpoint at step 100.  All splits use the common CAD verifier and cap $\max(3N,24)$.  Alignment intervals follow Table~\ref{tab:cad-panel-settings}.

\paragraph{CAD protocol for anytime work and budget distributions.}
The CAD work panel uses 48 ASAP training assemblies, disjoint from earlier CAD evaluation panels in both assembly identity and mesh content.  A deterministic pseudorandom order based only on assembly identity is fixed before evaluation and selects the parts quotas in Table~\ref{tab:cad-panel-settings}.  Evaluation uses the sensor seeds, lifts, and proposal cap in Table~\ref{tab:cad-panel-settings}, with noise of five degrees applied to the process axis and the common CAD verifier.  Work $W$ is the cumulative number of sampled sweep positions checked during proposal verification, with 1--96 positions per call.  At each position, assembly poses are reset and MuJoCo evaluates contact distances.  Initial setup of the contact baseline and final sequence replay are excluded; a cached verification is charged its original sweep count.  At work budget factor $b$, success requires a valid completion with $W\le\lceil 96bN\rceil$.

The work budget grid is $B=(1.25,1.5,2,2.5,3,4)$ full sweep equivalents per part.  The three predeclared interruption distributions on this ordered grid are
\[
\begin{aligned}
 \mu_{\rm tight}&=(.35,.25,.18,.12,.07,.03),\\
 \mu_{\rm uniform}&=(1/6,1/6,1/6,1/6,1/6,1/6),\\
 \mu_{\rm relaxed}&=(.03,.07,.12,.18,.25,.35).
\end{aligned}
\]
For each distribution, $\mathrm{AUC}_\mu=\sum_{b\in B}\mu(b)F(b)$.  The CAD work AUC reported in Table~\ref{tab:main} uses uniform weights; tight and relaxed distributions provide supplementary budget weightings.  Capped normalized work cost is $W/(96\cdot4N)$ for a valid completion and 1 otherwise, and the CAD work ``Cost reduction'' entry is the paired decrease from Static to \textup{SymBuild}.  Intervals follow Table~\ref{tab:cad-panel-settings}.

The supplementary diagnostic budgets in Table~\ref{tab:cad-work-fine} are $\{1.00,1.025,1.05,1.10,1.15,1.20,1.25\}$, chosen to resolve the observed work range for valid completions $1.000$--$1.161$ and include the original grid's starting point.  This analysis reuses all $48\times2\times4=384$ paired \textup{SymBuild}/Static conditions, preserving the verifier, rankings, sensor seeds, lifts, $4N$ cap on proposal calls, and the completion criterion $W\le\lceil96bN\rceil$ above; final replay remains excluded from work.  At each budget, the eight conditions are averaged within each assembly before 10,000 paired bootstrap draws with seed 23101.  The original AUC resampling indices are shared across diagnostic budgets, reproducing its interval at $b=1.25$.

\FloatBarrier
\subsection{Mini-Programs}
\label{app:program-protocol}
\paragraph{Program generation.}
Each proposed Mini-Program first draws its leaf count uniformly from the requested inclusive range.  A full binary expression tree is then generated recursively: for a subtree with more than one leaf, the number of leaves assigned to the left child is uniform on $\{1,\ldots,L-1\}$, and every internal operator is independently sampled uniformly from XOR, AND, and OR.  The two symmetric modules receive independent literal vectors of length $L$ whose entries are fair binary bits, and the final root operator is sampled uniformly from XOR and equality.  If the two module outputs are $v_0,v_1\in\{0,1\}$ and the root choice is indexed by $h\in\{0,1\}$, the generation label, which takes six possible values, is $3h+v_0+v_1$.  For a requested population of $C$ objects, the generator fills these six labels as evenly as possible; $C=120$ therefore gives exactly 20 programs per label.  Duplicate canonical programs are rejected with the two module orders identified under terminal swap; samples for labels whose quotas are full are skipped, and the accepted population is finally shuffled.  Leaf counts, tree shapes, internal operators, and literal patterns follow the sampling rules above.

\paragraph{Relation training and confirmation population.}
The frozen checkpoint uses 240 training and 60 validation objects with 3--5 leaves, disjoint from all development and confirmation objects.  Training and validation use base data seed 20261222, with generator offsets $+801$ and $+802$, respectively.  The relation network is trained with the settings in Table~\ref{tab:program-training}.  Validation macro F1 selects the checkpoint and the precedence, commutativity, and orbit thresholds.  The selected checkpoint and thresholds are used for confirmation.  Two frozen data seeds, 20270117 and 20270129, each generate 120 ID programs with 3--5 leaves using the data seed plus 901 as the generator seed and 120 length OOD programs with 6--8 leaves using the data seed plus 902 as the generator seed, giving 240 objects per split.  Object identities are pairwise disjoint across seed and split combinations and from earlier Mini-Program experiments.  The deployment interface retains each direct reference independently with probability $.35$ in one randomly chosen anchor module, yielding the observable partial reference graph used at confirmation.

\begin{table}[h]
\centering
\caption{\textbf{Mini-Program relation training.} Validation macro F1 is evaluated every 100 steps to select the checkpoint and relation thresholds. Data populations and generator seeds are specified in the preceding protocol.}
\label{tab:program-training}
\footnotesize
\begin{tabular}{ll}
\toprule
Setting & Value\\
\midrule
Network & Width 64 with three message-passing layers\\
Optimizer & AdamW\\
Training steps & 800\\
Training seed & 211\\
Learning rate & $.002$\\
Batch size & 32\\
Gradient clipping & 1\\
Validation interval & 100 steps\\
Selected thresholds & Precedence $.75$, commutativity $.20$, orbit $.10$\\
\bottomrule
\end{tabular}
\end{table}

\paragraph{Proposal ranks and state updates.}
Each program presents the same unordered event pool to every method.  Events are indexed by module and expression node, with a fixed final root; candidates are unfinished events.  Swapping modules exchanges corresponding event indices and fixes the root.  Precedence matrices have predecessors as rows and successors as columns.  Let $O$ be the sparse observed reference graph and $S$ the cross-module correspondence matrix, with zero diagonal and root row and column.  The observed expert adjoins references propagated by $(SOS)>0$, adds constraints from non-root events to the root, removes self edges, and takes transitive closure.  The frozen relation model supplies precedence probabilities; their elementwise maximum with the aligned observed graph defines the fixed process graph.  All matrices share the seeded event presentation order.  At state $s$ with remaining event mask $R_s$, either graph $G$ assigns candidate $a$ the residual predecessor mass
\[
  z_G(s,a)=\sum_{u\in R_s} G_{ua},
\]
with lower values preferred; equal scores receive their average weak rank.  The process rank $r_T$ uses the graph combining learned and observed relations and the state rank $r_X$ uses the transported observed graph.  \textup{SymBuild} recomputes both current ranks after every accepted event.  Static computes the same two ranks once at the empty prefix and subsequently only removes completed events.  The exact proposal key for meet is $(\min(r_T,r_X),\,r_T+r_X,\,\max(r_T,r_X),\,\pi(a))$, where $\pi$ is a deterministic presentation order fixed by the seed; the Process-only and State-only controls use $r_T$ and $r_X$, respectively.  The permuted control applies a fixed random permutation to all non-root event identities, conjugates only the learned precedence matrix, and keeps the observed expert aligned.  For data seed $d$ and split index $j\in\{0,1\}$ for ID and length OOD, reference sampling uses seed $d+1900+100000j$.  For object index $i$ within the group defined by the data seed and split, presentation order and planning ties use $d+100000j+1009i$.  Every proposal is accepted exactly when the hidden ground truth precedence graph contains no unfinished predecessor of that event.  The hidden directed acyclic graph (DAG) preserves a complete topological suffix after each accepted event.

\paragraph{Mini-Program anytime metric.}
The maximum budget is $2n$ verifier calls for $n$ events, with the predeclared grid $q/n\in\{1,1.25,1.5,1.75,2\}$.  Program anytime AUC is the unweighted mean of the five completion indicators on this grid, i.e., the Section~\ref{sec:anytime} AUC definition with uniform mass on these five budgets.  Capped normalized cost is $q/(2n)$ for a successful plan and $1$ for an unsuccessful plan; the Program ``Cost reduction'' entries in Table~\ref{tab:main} report the paired Static minus \textup{SymBuild} reduction in this quantity.  Intervals use 10,000 bootstrap draws paired by object (seed 24101), and the two frozen data seeds are also reported separately as replication groups.

\FloatBarrier
\subsection{Exact-Fill Packing}
\label{app:packing-protocol}
\paragraph{Instance generation and size shift.}
A packing action is $a=(i,b)$, assigning an unassigned item $i$ of weight $w_i$ to bin $b$ of capacity 100.  Each latent group contains exactly three items whose weights sum to 100.  The largest weight $w_1$ is sampled uniformly from the integers 45--60, and a second weight $w_2$ from 12--30.  The third weight is set to the residual $100-w_1-w_2$; sampling is repeated until the residual lies in 12--40 and at least two of the three weights differ.  Groups and items are permuted, and a random permutation maps latent groups to terminal bin identities.  One visible anchor per group (the heaviest item, with item index breaking weight ties) is preassigned to its transported bin, so the anchors expose a complete bijection $\tau$ from groups to bins.  Among the $n$ remaining items, $\max(2,\operatorname{round}(.25n))$ reference group labels are selected without replacement and cyclically shifted, implementing the 25\% prior corruption condition.  With data seeds 20271103 and 20271117, each seed contributes 120 ID instances with 4--6 bins and 120 size OOD instances with 7--8 bins.  Hence ID instances contain 12--18 total items and 8--12 initially unassigned decision items, while size OOD instances contain 21--24 total items and 14--16 initially unassigned decision items.

\paragraph{Transported prior, proposal ranks, and state updates.}
Let $g_i$ be the observed reference group label.  The fixed transported structure is $i\mapsto\tau(g_i)$: the group-to-bin bijection acts on reference assignment labels while item identities remain fixed.  At state $s$, $A_s$ contains exactly the assignments respecting current capacities.  Let $L_b(s)$ be current load and $c_i(s)=|\{b:(i,b)\in A_s\}|$.  For a lexicographic key $\kappa$, $\operatorname{ordnorm}_{A_s}(\kappa)$ sorts by $(\kappa(a),a)$ and maps the zero-based position to $[0,1]$ by division by $\max(|A_s|-1,1)$.  The transported and residual ranks are
\[
\begin{aligned}
 r_T(i,b)&=\operatorname{ordnorm}_{A_s}\!\left(\mathbf{1}[b\neq\tau(g_i)],\,g_i,\,-w_i,\,b\right),\\
 r_X(i,b;s)&=\operatorname{ordnorm}_{A_s}\!\left(c_i(s),\,-w_i,\,100-L_b(s)-w_i,\,b\right).
\end{aligned}
\]
\textup{SymBuild} queries actions by $(\min(r_T,r_X),r_T,r_X,a)$.  Both variants recompute $r_T$ on current $A_s$; refresh also recomputes $r_X$ after acceptance, while Static retains initial residual ranks for surviving actions.  The process-only control uses $r_T$.  The best fit control orders by $(-w_i,100-L_b(s)-w_i,b)$.  Since $c_i(s)$ is nonincreasing in $w_i$, best fit and $r_X$ induce the same ordering under the shared tie-break.  Permuted transport cyclically shifts the recovered group-to-bin map.

\paragraph{Verification, budgets, and statistics.}
The verifier rejects capacity overflow and sum mismatch, then accepts only when memoized exact branch-and-bound proves that remaining weights exactly fill residual capacities.  Visible anchors recover the group-to-bin correspondence while corrupted non-anchor labels affect the prior.  The call grid is $q/n\in\{1,1.125,1.25,1.5,2,3\}$ for $n$ initially unassigned items.  Packing AUC uniformly averages these budgets; capped call cost is $q/(3n)$ on success and 1 on failure, and Table~\ref{tab:main} reports paired Static minus \textup{SymBuild} reductions.  Intervals use 10,000 paired instance bootstrap draws with base seed 26101; deterministic offsets index split, contrast, metric, budget, and replication for each data seed.

\FloatBarrier
\subsection{IKEA Observed Process Topology}
\label{app:ikea-protocol}
\paragraph{Terminal representation.}
We use the IKEA Video Manuals dataset from \citet{liu2024ikea}, which provides aligned furniture parts, manuals, and substeps from real assembly videos.  The resulting dataset contains 36 terminal families and 97 parsed real videos.  The 267 referenced OBJ parts provide the identities and geometry used to construct the terminal representation.  Terminal correspondence is computed from the exact part orbits of an unordered graph of manual incidence under automorphisms that preserve color, with witness automorphisms supplying the transport maps used for candidate closure.  For each OBJ part, vertices are centered and divided by their root-mean-square (RMS) radius.  The part color is determined by the normalized covariance eigenvalues in descending order, quantiles of radial distance at $0,.1,.25,.5,.75,.9,1$, and the logarithm of the vertex count to base two; eigenvalues and radial quantiles are rounded to three decimals and the statistic for vertex count to two decimals.  A mesh with zero radius receives a separate degenerate color indexed by its vertex count.  The resulting color is invariant under rigid transformations and independent of pose and process order.  Each official manual connection is represented by one connection node joined to two side nodes of the same color, each connected to the corresponding endpoint part.  The terminal graph contains only the unordered incidence structure among parts and the part colors defined above, which are invariant under rigid transformations.  Thirteen families contain nontrivial terminal orbits.

\paragraph{Event selection and candidate construction.}
The video parser admits only nonempty video IDs and substeps with a recorded start time and at least one nonempty part component.  Within each video, usable events are sorted by start time and source order; consecutive exact repeats with the same action identity and start time are deduplicated.  A predictive state is formed between successive usable events only when their start times differ.  These deterministic parsing rules produce the reported 97 videos and 1,022 evaluable states; the latter span 35 families because one of the 36 terminal families contributes no evaluable transition.  Each state is the realized video prefix up to the current substep.  Its candidate set is the deduplicated union of all remaining observed actions and their variants induced by terminal transport maps, and the positive label is the actually observed next action.  Of the 1,022 states, 191 states in 13 families exhibit terminal ambiguity, meaning that the next action admits a nontrivial transported alternative.

\paragraph{Predictors and statistical evaluation.}
All compared predictors use the same candidates, five folds with families held out for out-of-fold evaluation, 4,096-dimensional feature hashing, logistic regression, seed 20260802, and $C\in\{.03,.1,.3,1,3\}$ selected by the predeclared rule based on macro mean reciprocal rank (MRR) on the validation data.  The prefix model uses candidate features and features of the realized prefix; the static orbit model adds fixed statistics of the terminal orbit; the section model refines orbit members using the prefix group in which each is first seen and its age; and the shuffled section control preserves terminal structure and prefix length while permuting which process member is assigned to the realized trajectory.  Top-1 and MRR rank the true next action among candidates, with paired bootstrap intervals at the family level from 10,000 draws.

\paragraph{IKEA quantities in Table~\ref{tab:ablation}(c).}
``Process structure'' is the Top-1 gap between the section and prefix models on all 191 states with terminal ambiguity.  ``Action core'' uses the predeclared core subset (125 states, 9 families), in which terminal ambiguity is present, the realized prefix breaks every relevant transport, and the terminal has 8--20 part nodes.  On this subset, prefix and static orbit have identical Top-1 accuracy, so the reported value is the section model's gain over either baseline.  ``Hybrid'' applies the section model on states with terminal ambiguity and the prefix model elsewhere; its reported value is the Top-1 gap against prefix over all 1,022 evaluable states.  Table~\ref{tab:ablation}(c) reports the paired gaps and intervals; the shuffled section control preserves terminal structure while changing which member of the terminal orbit the realized prefix selects.

\subsection{Alternative Aggregation and Matched Controls}
\label{app:aggregation-protocol}
\paragraph{Product rule and multi-heuristic alternatives.}
The SayCanPay-adapted product rule is parameter-free and uses the same process and state score channels as the matched primary comparison, sorting ascending by
\[
 K_{\rm prod}(a)=\bigl((1+r_T(a))(1+r_X(a)),\;r_T(a)+r_X(a),\;\max(r_T(a),r_X(a)),\;\text{tie}(a)\bigr).
\]
For CAD, the two raw channel scores are independently converted to SciPy average ranks shifted to start at zero; the first three key fields are encoded as product $+10^{-9}$ sum $+10^{-12}$ max, followed by the stable action order.  Mini-Programs use the average weak ranks and deterministic presentation order defined above.  Packing uses the normalized ranks above and the action tuple as its final tie-break.  The MH key columns in Appendix Table~\ref{tab:aggregation-full}(b) use a separate parameter-free key on the same interface
\[
 K_{\rm MH}(a)=\bigl(\min(2r_T(a),2r_X(a)+1),\;r_T(a)+r_X(a),\;\max(r_T(a),r_X(a)),\;\text{tie}(a)\bigr),
\]
while Appendix~\ref{app:search-controls} separately evaluates the MHA*-style $1{:}2$ scheduler selected on the development set.  For CAD, the static versions cache the initial combined score; for Mini-Programs they cache both initial weak ranks and then remove completed events; for exact-fill packing they retain the initial residual rank on surviving actions while restricting the fixed transported prior to the current candidate set.  Refresh versions recompute the score or rank derived from the current state after each accepted transition.  Candidate sets, verifier semantics, object splits, and interruption budgets match the primary panels using rank meet.

\paragraph{Matched controls.}
Refresh and static pairs share the instance, information channels, deterministic tie-break, proposal cap, and verifier.  CAD controls using a single information channel are PCS-ICP and ASAP-style outsideness.  Mini-Program process and state controls use combined learned and observed process relations and transported observed relations, respectively.  Exact-fill packing controls are the process prior and best fit.  Mini-Programs and exact-fill packing use permuted transport, while the CAD panel that uses a learned residual and a disjoint set of objects supplies the alignment intervention.  Confirmation objects have identities disjoint from development and from one another.  The comparison with the physics oracle in the CAD mechanism panel serves a diagnostic role.

\subsection{Implementation Sources and Environment}
\paragraph{External implementation provenance.}
The CAD source assemblies and outside-in geometric control are grounded in ASAP \citep{tian2024asap}; CAD verification uses MuJoCo \citep{todorov2012mujoco}.  The product-rule comparison adapts the multiplicative scoring rule of SayCanPay \citep{hazra2024saycanpay} using the exact joint key defined above.  GRN experiments use the official GRN checkpoint and Panda OOD data \citep{bouhsain2025grn}, and the population-based search comparison follows the CDGS search design of \citet{mishra2026cdgs}.  Table~\ref{tab:open-code-compare} summarizes how the public SayCanPay, GRN, and PNO implementations enter matched comparisons or task-level context; the CDGS adaptation is specified separately in Appendix~\ref{app:grn-search}.  The implementations in the table use fixed repository revisions.

\paragraph{Software and hardware.}
Table~\ref{tab:environment} specifies the computing environment.  Learned relation and GRN scoring use CUDA.  Discrete verifiers, bootstrap and permutation statistics, and table aggregation run on CPU.

\begin{table}[!ht]
\centering
\caption{\textbf{Software and hardware used for the reported experiments.}}
\label{tab:environment}
\footnotesize
\begin{tabular}{p{.18\textwidth}p{.74\textwidth}}
\toprule
Component & Configuration\\
\midrule
Operating system & 64-bit Windows 11\\
CPU and RAM & AMD Ryzen 7 8845H with 8 cores and 16 threads, 31.3\,GiB RAM\\
GPU and driver & NVIDIA GeForce RTX 4060 Laptop with 8,188\,MiB memory and driver 572.83\\
Python and PyTorch & Python 3.8.20 and PyTorch 2.4.1 with CUDA 12.4\\
Libraries & NumPy 1.24.4, SciPy 1.10.1 and Matplotlib 3.7.2\\
\bottomrule
\end{tabular}
\end{table}

\end{document}

%% file: math_commands.tex
\usepackage{amsmath,amsfonts,bm}

\def\eqref#1{equation~\ref{#1}}
\def\Eqref#1{Equation~\ref{#1}}

\def\1{\bm{1}}

\DeclareMathAlphabet{\mathsfit}{\encodingdefault}{\sfdefault}{m}{sl}
\SetMathAlphabet{\mathsfit}{bold}{\encodingdefault}{\sfdefault}{bx}{n}



%% file: approx_verification_appendix.tex
\subsubsection{Approximate Verification}

\begin{table}[h]
\centering
\caption{\textbf{Complete stochastic verifier robustness test.} True AUC uses hidden exact validity. FC denotes false certification. $\Delta$AUC is \textup{SymBuild} minus Static, and FC pairs follow the same order. Higher values are better for true AUC and $\Delta$AUC ($\uparrow$); lower values are better for FC ($\downarrow$). Values are averaged over 20 noise seeds within each 240-object split before paired bootstrap.}
\label{tab:verifier-noise}
\scriptsize
\setlength{\tabcolsep}{3.0pt}
\begin{tabular}{lrrrrr}
\toprule
$p$ & \textup{SymBuild} true AUC & Static true AUC & $\Delta$AUC $[95\%\,\mathrm{CI}]$ & FC (\textup{SymBuild} / Static) & Seeds (objects)\\
\midrule
\multicolumn{6}{l}{\textbf{False positive -- ID}}\\
.01 & \textbf{83.43} & \underline{69.81} & $+13.62$ $[10.53,16.76]$ & \textbf{2.60}/\underline{5.38} & 20 (240)\\
.025 & \textbf{81.36} & \underline{66.08} & $+15.28$ $[11.92,18.73]$ & \textbf{5.79}/\underline{12.21} & 20 (240)\\
.05 & \textbf{78.27} & \underline{60.80} & $+17.47$ $[13.70,21.25]$ & \textbf{10.67}/\underline{21.56} & 20 (240)\\
.10 & \textbf{72.72} & \underline{52.09} & $+20.62$ $[16.03,25.20]$ & \textbf{18.92}/\underline{36.46} & 20 (240)\\
\midrule
\multicolumn{6}{l}{\textbf{False positive -- Length OOD}}\\
.01 & \textbf{60.75} & \underline{38.81} & $+21.94$ $[18.68,25.26]$ & \textbf{8.10}/\underline{12.42} & 20 (240)\\
.025 & \textbf{54.63} & \underline{33.24} & $+21.39$ $[18.10,24.63]$ & \textbf{19.88}/\underline{28.17} & 20 (240)\\
.05 & \textbf{46.78} & \underline{26.30} & $+20.48$ $[17.18,23.76]$ & \textbf{34.12}/\underline{48.12} & 20 (240)\\
.10 & \textbf{35.62} & \underline{17.72} & $+17.90$ $[14.61,21.23]$ & \textbf{52.85}/\underline{70.48} & 20 (240)\\
\midrule
\multicolumn{6}{l}{\textbf{False negative -- ID}}\\
.01 & \textbf{81.04} & \underline{69.64} & $+11.40$ $[8.59,14.16]$ & 0/0 & 20 (240)\\
.025 & \textbf{74.96} & \underline{65.01} & $+9.95$ $[7.39,12.57]$ & 0/0 & 20 (240)\\
.05 & \textbf{66.97} & \underline{58.33} & $+8.65$ $[6.25,11.13]$ & 0/0 & 20 (240)\\
.10 & \textbf{52.88} & \underline{45.97} & $+6.92$ $[4.84,9.06]$ & 0/0 & 20 (240)\\
\midrule
\multicolumn{6}{l}{\textbf{False negative -- Length OOD}}\\
.01 & \textbf{61.28} & \underline{40.74} & $+20.54$ $[17.35,23.73]$ & 0/0 & 20 (240)\\
.025 & \textbf{55.97} & \underline{38.09} & $+17.88$ $[14.94,20.85]$ & 0/0 & 20 (240)\\
.05 & \textbf{48.57} & \underline{33.56} & $+15.01$ $[12.36,17.75]$ & 0/0 & 20 (240)\\
.10 & \textbf{36.38} & \underline{24.98} & $+11.40$ $[9.16,13.60]$ & 0/0 & 20 (240)\\
\bottomrule
\end{tabular}
\end{table}

On the same 480 confirmation objects, each combination of object, completed set, candidate action, regime, and noise seed receives a deterministic noise draw shared across paired methods.  For $p\in\{.01,.025,.05,.10\}$, false positives accept invalid actions and mark them complete, while false negatives reject valid actions and leave the state unchanged, each with probability $p$; all other responses match the exact verifier.  A hidden flag tracks true prefix validity.  We average over noise seeds 0--19 for each object and condition before 10,000 paired bootstrap draws.  Bootstrap sampling uses base seed 61000 with deterministic offsets for split, noise regime, noise level, and metric.  Completion marks all events complete; true success additionally requires every accepted action to be valid; false certification is completion without true success.

\FloatBarrier

%% file: cad_work_fine_appendix.tex
\paragraph{CAD work at finer budget resolution.}
Supplementary diagnostic budgets resolve the completion thresholds before the original CAD work grid begins.  Table~\ref{tab:cad-work-fine} gives verified success over the same 384 paired evaluation conditions at this finer resolution.

\begin{table}[H]
\centering
\caption{\textbf{Finer-resolution CAD work diagnostic.}  Work budgets are full sweep equivalents per part.  Entries reuse the original 48 assemblies, two sensor seeds, four lifts, and $4N$ cap on proposal calls.  Success percentages are measured over all evaluation conditions; differences use 10,000 paired bootstrap draws over assemblies after averaging sensor and lift conditions.}
\label{tab:cad-work-fine}
\small
\setlength{\tabcolsep}{5.0pt}
\begin{tabular}{lrrl}
\toprule
\shortstack{Work budget $b$\\(sweeps per part)} & \shortstack{\textup{SymBuild}\\success (\%)} & \shortstack{Static\\success (\%)} & $\Delta$ success (pp) [paired 95\% CI]\\
\midrule
1.000 & \underline{37.24} & \textbf{38.54} & $-1.30\,[-12.76,10.16]$\\
1.025 & \textbf{77.34} & \underline{68.23} & $+9.11\,[-1.04,20.05]$\\
1.050 & \textbf{88.80} & \underline{76.04} & $+12.76\,[3.13,23.44]$\\
1.100 & \textbf{97.14} & \underline{84.64} & $+12.50\,[3.65,22.40]$\\
1.150 & \textbf{98.70} & \underline{90.63} & $+8.07\,[2.34,15.36]$\\
1.200 & \textbf{98.70} & \underline{91.93} & $+6.77\,[1.30,14.06]$\\
1.250 (original start) & \textbf{98.70} & \underline{91.93} & $+6.77\,[1.30,14.06]$\\
\bottomrule
\end{tabular}
\end{table}

At $b=1.05$ and $1.10$, \textup{SymBuild} raises verified success by $12.76$ and $12.50$ points, respectively.  The two methods reach their completion plateaus by $b=1.20$, with \textup{SymBuild} completing 26 additional paired conditions.  These diagnostics expose both higher completion probabilities at smaller budgets and greater completion coverage under the fixed proposal cap.  On the original grid, each method's success curve is constant, so its AUC equals its plateau success and the tight, uniform, and relaxed interruption distributions give the same value.

%% file: aggregation_transfer_appendix.tex
\paragraph{Development control for rank aggregation.}
The CAD development comparison favors rank meet in both success and verifier cost.  The panel contains 60 objects with four fixed $C_4$ lifts each and holds both rank channels and the verifier fixed.  Rank meet achieves $86.25\%$ success at cost $1.547$, compared with $80.00\%$ at $1.752$ for Borda count and $79.17\%$ at $1.820$ for rank join, the pointwise maximum of the two ranks.  This joint improvement supports the choice of rank meet for confirmation (Section~\ref{sec:search-controls}).

\paragraph{Packing development control.}
A disjoint panel of 80 instances uses seeds 314159 and 271828, with 40 instances per seed and 4--6 bins.  It uses capacity 100, 25\% prior corruption, the exact extension verifier, and the confirmation budget grid.  Rank meet, Borda count, and rank join are compared using 10,000 bootstrap draws paired by instance with base seed 25101; rank meet is then used for confirmation.

\paragraph{Weighted sum selection.}
A disjoint 160-object Mini-Program development set uses seeds 20270203 and 20270211, with 40 objects per seed and split and identities disjoint from confirmation and earlier experiments.  It evaluates weighted rank sums $w r_T+(1-w)r_X$ for $w\in\{.25,.50,.75\}$.  The mean of ID and OOD development AUC selects $w=.25$, which is frozen for the 480-object confirmation panel.  Refreshed and static variants share checkpoint, candidates, verifier, budgets, tie-breaking, and the bootstrap protocol at the object level.

\begin{table}[h]
\centering
\caption{\textbf{Alternative aggregation confirmation.}  Panel (a) compares rank meet with the weighted sum selected on development data for Mini-Programs; AUC, $\Delta$AUC, and final success are better when higher ($\uparrow$), while cost is better when lower ($\downarrow$).  Panel (b) reports the SayCanPay-adapted product rule and a parameter-free multi-heuristic key (MH key) on all six construction panels. Each cell reports AUC followed by final success, both as percentages.}
\label{tab:aggregation-full}
\scriptsize
\setlength{\tabcolsep}{2.7pt}
\resizebox{\textwidth}{!}{%
\begin{tabular}{llrrrrll}
\toprule
\multicolumn{8}{l}{\textbf{(a) Weighted aggregation selected on development data}}\\
Split & Aggregator & Refresh AUC & Static AUC & $\Delta$AUC $[95\%\,\mathrm{CI}]$ & Final (refresh / static) & Cost (refresh / static) & Selection\\
\midrule
ID & Rank meet & \textbf{85.08} & \underline{72.67} & $+12.42$ $[9.50,15.33]$ & 100/100 & \textbf{.5719}/\underline{.6456} & primary\\
ID & Weighted sum $w=.25$ & \textbf{85.17} & \underline{71.67} & $+13.50$ $[10.50,16.42]$ & 100/100 & \textbf{.5718}/\underline{.6530} & development\\
Length OOD & Rank meet & \textbf{64.67} & \underline{42.92} & $+21.75$ $[18.33,24.92]$ & \textbf{97.50}/\underline{83.33} & \textbf{.6706}/\underline{.8057} & primary\\
Length OOD & Weighted sum $w=.25$ & \textbf{65.83} & \underline{41.67} & $+24.17$ $[20.75,27.50]$ & \textbf{97.50}/\underline{82.08} & \textbf{.6636}/\underline{.8132} & development\\
\bottomrule
\end{tabular}}\par\nointerlineskip
\setlength{\tabcolsep}{3.0pt}
\begin{tabular*}{\textwidth}{@{\extracolsep{\fill}}lrrrr@{}}
\multicolumn{5}{l}{\textbf{(b) Alternative aggregators in the three construction domains}}\\
Panel & Product refresh & Product static & MH key refresh & MH key static\\
\midrule
CAD calls & 82.38/95.49 & 79.08/90.10 & 81.63/94.44 & 78.76/89.76\\
CAD work & 98.18/98.18 & 93.75/93.75 & 98.70/98.70 & 91.93/91.93\\
Program ID & 85.08/100 & 72.67/100 & 85.08/100 & 72.83/100\\
Program OOD & 65.50/97.50 & 42.92/83.75 & 64.33/97.50 & 42.83/82.50\\
Packing ID & 72.29/100 & 65.83/100 & 50.21/100 & 50.21/100\\
Packing OOD & 47.92/98.33 & 41.39/99.17 & 28.40/93.75 & 28.40/93.75\\
\bottomrule
\end{tabular*}
\end{table}

\FloatBarrier

%% file: search_controls_appendix.tex
\subsection{Alternative Planning Methods}
\label{app:search-controls}

\paragraph{Common interface.}
At the search interface, $r_T^s$ and $r_X^s$ support multiple proposal methods: multi-heuristic scheduling keeps separate queues, partial-order causal-link (POCL)-style scheduling \citep{mcallester1991systematic} constrains an agenda, LazySP-style selection performs prefix search without verifier calls, and rank meet compiles the prefixes into one order.  Section~\ref{sec:search-controls} tests these framework instantiations.

These controls use the same 480 Mini-Programs, frozen relation model, candidates, verifier, budgets, and deterministic tie order as the primary panel.  Each data group contains 120 ID objects (3--5 leaves) and 120 length OOD objects (6--8 leaves).  The cap is $2n$ verifier calls and anytime AUC averages success at $q/n\in\{1,1.25,1.5,1.75,2\}$.  Paired intervals use 10,000 bootstrap draws paired by object; search and canonicalization comparisons use base seed 28101, while comparisons between refreshed and fixed schedulers use base seed 28201.  Deterministic offsets index split, contrast, metric, and budget.

\paragraph{MHA*-style scheduling and selection.}
A disjoint 160-object development set tests service ratios between the process and state queues $1{:}1,1{:}2,2{:}1,1{:}3,3{:}1$ and selects $1{:}2$ by mean ID/OOD AUC.  For a ratio $p{:}s$, each cycle emits the next $p$ unseen candidates ordered by process rank and then the next $s$ candidates ordered by state rank, skipping duplicates; the selected ratio is frozen before confirmation.

The transported process rank and state residual rank define the two queues.  The refreshed variant rebuilds both after every accepted event; the fixed counterpart filters the initial queues as events are completed without recomputing their ranks.

\paragraph{Canonical slice control.}
The exact terminal correspondence defines an orbit under identity and swap.  For each program, the observable serialization of both representatives is formed from event types, numeric attributes, and observed edges; the lexicographically smaller representative is selected.  Its canonical event positions provide the final tie resolution for the Static order produced by rank meet.  Thus the control uses exact terminal canonicalization to reduce ambiguity in orbit labels while retaining the same observable information and verifier access as the primary method.  State evidence is held at its initial value, isolating representative selection from refresh after transitions.

\paragraph{POCL-style partial-order agenda.}
An observable acyclic partial order is constructed from transported observed edges and learned edges of probability at least $.5$; any edge that would create a cycle is dropped.  At each state, candidates with fewer remaining predecessors are served first; state and process ranks, followed by the shared deterministic tie order, resolve agenda ties.  The refreshed variant recomputes those ranks after every accepted event.  Its fixed counterpart updates which predecessors remain but retains the initial state and process ranks for tie resolution.

\paragraph{Weighted A* with LazySP-style replanning.}
A bounded best-first search proposes short candidate paths from the observable projected partial order and state rank using prefix expansion without verifier calls.  Search width is 32, horizon is 4, and the weighted best-first coefficient is 2.  The selected path contributes its first action to the shared verifier.  Rejection invalidates that state-action choice and triggers replanning; acceptance advances the state and restarts planning from the successor.

\begin{table}[h]
\centering
\caption{\textbf{Full comparison of search and canonicalization under matched interfaces.}  AUC and final success are percentages.  Selector time denotes mean Python proposal time per object; expansions count prefix expansions performed without verifier calls.  All rows share the same 480-object population, observable information, verifier, and $2n$ cap.}
\label{tab:search-full}
\scriptsize
\setlength{\tabcolsep}{2.8pt}
\resizebox{\textwidth}{!}{%
\begin{tabular}{lrrrrrrrr}
\toprule
& \multicolumn{4}{c}{ID} & \multicolumn{4}{c}{Length OOD}\\
\cmidrule(lr){2-5}\cmidrule(lr){6-9}
Method & AUC $\uparrow$ & Success@$2n$ $\uparrow$ & Selector ms $\downarrow$ & Expansions $\downarrow$ & AUC $\uparrow$ & Success@$2n$ $\uparrow$ & Selector ms $\downarrow$ & Expansions $\downarrow$\\
\midrule
\textup{SymBuild} & 85.08 & 100.00 & 5.5 & 0 & 64.67 & 97.50 & 11.2 & 0\\
Static & 72.67 & 100.00 & 2.0 & 0 & 42.92 & 83.33 & 3.2 & 0\\
POCL-style refresh & 85.67 & 100.00 & 9.1 & 0 & 66.33 & 97.92 & 21.8 & 0\\
POCL-style fixed & 84.58 & 100.00 & 6.1 & 0 & 62.50 & 96.67 & 13.5 & 0\\
Canonical slice with static meet & 72.50 & 100.00 & 8.3 & 0 & 42.92 & 82.92 & 19.9 & 0\\
MHA*-style $1{:}2$ refresh & 85.92 & 100.00 & 8.4 & 0 & 65.83 & 97.08 & 20.4 & 0\\
MHA*-style $1{:}2$ fixed & 86.33 & 100.00 & 5.1 & 0 & 60.83 & 95.00 & 11.9 & 0\\
Weighted A* with LazySP-style replanning & 80.67 & 99.58 & 94.2 & 391.7 & 60.42 & 96.25 & 193.3 & 609.6\\
\bottomrule
\end{tabular}}
\end{table}

On length OOD, refresh reduces capped normalized verifier cost by $2.35$ points for POCL-style ($95\%$ CI $[1.81,2.91]$) and $3.26$ for MHA*-style ($[2.11,4.42]$).  The MHA*-style costs are $.5652$ for refresh and $.5641$ for the fixed variant on ID, and $.6635$ and $.6962$, respectively, on length OOD.

%% file: grn_ordinal_appendix.tex
\paragraph{Ordinal fusion with learned state scores.}
The ordinal methods use the same 1,135 reachable targets, three OOD scales, candidate sets, fixed verifier, and $5N$ cap as the learned scoring methods above.  Here $N$ counts initially movable objects.  The process structure is the scene's annotated obstruction and support graph, aligned by the identity correspondence on the shared object indices.  For each full remaining candidate set $A_s$, negative process and GRN utilities are separately converted to one-based average weak ranks $r_T^s$ and $r_X^s$, with exact equality defining ties.  Rank meet orders candidates lexicographically by
\[
\bigl(\min(r_T^s,r_X^s),\ r_T^s+r_X^s,\ \max(r_T^s,r_X^s),\ \pi(a)\bigr).
\]
Borda and rank product use the respective keys
\[
\begin{aligned}
 &(r_T^s+r_X^s,\max(r_T^s,r_X^s),\pi(a)),\\
 &(r_T^sr_X^s,r_T^s+r_X^s,\max(r_T^s,r_X^s),\pi(a)).
\end{aligned}
\]
These parameter-free rules provide three ordinal realizations using identical information.  Static retains both initial raw channels, filters removed objects, and reconstructs ranks on the current full candidate set.  Refresh recomputes raw utilities after each accepted removal before constructing ranks.  Rejection masks the queried object--grasp pair while retaining the current state's scores and ranks.  The process-only controls follow the corresponding raw-score caching and refresh rules.

\paragraph{Matched evaluation and inference.}
The six selectors are evaluated in paired Static and refresh variants with seeds 0--4, giving 645, 2,160, and 2,870 conditions per selector and variant on OOD10, OOD15, and OOD20, respectively, and 68,100 runs in total.  Candidate priorities are sampled once for each scene, target, and seed and shared across methods; object and grasp indices resolve remaining ties.  The official checkpoint is fixed, and the ordinal rules require no additional training or parameter selection.  Bootstrap and sign permutation inference use the mean paired difference across targets after averaging the five seeds within each target.  With random seed 20260809, bootstrap inference forms 95\% percentile intervals from 20,000 resamples of entire scenes and averages all targets in each resample.  The two-sided sign permutation test uses 20,000 assignments, applying one shared sign to all target differences within each scene.  Table~\ref{tab:grn-direct-full} reports query scores, completion rates, and paired inference for all six selectors.  The tail query score reports the 95th percentile over conditions pooled across targets and seeds, using the same failure penalty as the mean score.

\begin{table}[h]
\centering
\caption{\textbf{Complete direct scoring and ordinal comparisons in GRN target removal.} Mean query scores and paired refresh minus static gaps use the common $5N$ cap.  Completion percentages list static first and refresh second.  The best and second-best query scores among all twelve conditions at each scale are bold and underlined; ties share emphasis.  The reported $p$ values come from a two-sided sign permutation test applied at the scene level.}
\label{tab:grn-direct-full}
\footnotesize
\setlength{\tabcolsep}{3pt}
\renewcommand{\arraystretch}{1.05}
\begin{tabular}{llrrlrr}
\toprule
Split & Selector & Static $\downarrow$ & Refresh $\downarrow$ & $\Delta$ $[95\%\,\mathrm{CI}]$ & Completion (\%) & $p$\\
\midrule
OOD10 & GRN & 5.457 & 5.132 & $-0.326\;[-0.743,0.066]$ & $100.00,\ 100.00$ & 0.17569\\
OOD10 & Process & 9.454 & 8.326 & $-1.129\;[-1.898,-0.464]$ & $97.98,\ 98.91$ & 0.00230\\
OOD10 & Combined & 3.349 & \textbf{2.597} & $-0.752\;[-1.279,-0.270]$ & $100.00,\ 100.00$ & 0.00645\\
OOD10 & Rank meet & 5.352 & 4.747 & $-0.605\;[-1.433,-0.025]$ & $99.22,\ 100.00$ & 0.09685\\
OOD10 & Borda & 3.073 & \underline{2.654} & $-0.419\;[-0.790,-0.118]$ & $99.22,\ 100.00$ & 0.02385\\
OOD10 & Rank product & 3.794 & 3.205 & $-0.589\;[-1.370,0.025]$ & $100.00,\ 100.00$ & 0.15639\\
\midrule
OOD15 & GRN & 8.213 & 6.734 & $-1.479\;[-2.053,-0.978]$ & $100.00,\ 100.00$ & 0.00005\\
OOD15 & Process & 15.492 & 12.301 & $-3.191\;[-4.202,-2.256]$ & $97.18,\ 98.84$ & 0.00005\\
OOD15 & Combined & 4.523 & \textbf{3.349} & $-1.174\;[-1.500,-0.865]$ & $100.00,\ 100.00$ & 0.00005\\
OOD15 & Rank meet & 7.097 & 6.127 & $-0.970\;[-1.472,-0.560]$ & $100.00,\ 100.00$ & 0.00005\\
OOD15 & Borda & 4.825 & \underline{3.576} & $-1.249\;[-1.650,-0.888]$ & $100.00,\ 100.00$ & 0.00005\\
OOD15 & Rank product & 5.575 & 4.356 & $-1.219\;[-1.614,-0.869]$ & $100.00,\ 100.00$ & 0.00005\\
\midrule
OOD20 & GRN & 11.027 & 8.028 & $-2.999\;[-4.372,-1.879]$ & $99.83,\ 100.00$ & 0.00005\\
OOD20 & Process & 21.912 & 15.755 & $-6.157\;[-8.229,-4.231]$ & $96.45,\ 98.85$ & 0.00005\\
OOD20 & Combined & 6.332 & \textbf{3.873} & $-2.459\;[-3.225,-1.774]$ & $100.00,\ 100.00$ & 0.00005\\
OOD20 & Rank meet & 9.319 & 7.608 & $-1.711\;[-2.608,-0.995]$ & $100.00,\ 100.00$ & 0.00005\\
OOD20 & Borda & 7.251 & \textbf{3.873} & $-3.378\;[-4.873,-2.115]$ & $99.13,\ 100.00$ & 0.00005\\
OOD20 & Rank product & 7.364 & \underline{4.898} & $-2.466\;[-3.376,-1.704]$ & $100.00,\ 100.00$ & 0.00005\\
\bottomrule
\end{tabular}
\end{table}

\paragraph{Verification tails and post-transition behavior.}
Matching the first action locates the benefit of refresh in subsequent decisions (Table~\ref{tab:grn-direct-diagnostic}).  Ordinal rank meet reduces mean query scores by $11.30\%$, $13.67\%$, and $18.36\%$ on OOD10, OOD15, and OOD20, respectively.  Its tail query score falls from 16 to 10 on OOD10, from 19 to 15 on OOD15, and from 25 to 17 on OOD20, showing that the benefit extends to difficult episodes.  Refresh also completes five OOD10 conditions left unfinished by Static.

The transported process channel complements learned feasibility: refreshed rank meet improves on GRN refresh by $0.606$ queries $[0.308,0.956]$ on OOD15 and $0.420$ queries $[0.191,0.698]$ on OOD20.  Combined and Borda combine target and grasp evidence additively; rank meet keeps candidates favored by either source early in the order.  Refresh gains across all six selectors support statewise refinement under these different aggregation choices, with Combined achieving the lowest mean query scores at all three scales.

\begin{table}[h]
\centering
\caption{\textbf{Execution diagnostics for the matched GRN selectors.} Scored states and logical GRN forwards list static first and refresh second, including initial scoring.  Sequence disagreement and unique completion counts use all conditions across the five seeds. Refresh only / Static only counts completions unique to each variant.  Every pair shares its first queried action.  Tail query score is the 95th-percentile query score and lists Static first and refresh second.}
\label{tab:grn-direct-diagnostic}
\footnotesize
\setlength{\tabcolsep}{3pt}
\renewcommand{\arraystretch}{1.05}
\begin{tabular}{llrrrcr}
\toprule
Split & Selector & \shortstack{Scored states\\(static, refresh)} & \shortstack{Logical GRN forwards\\(static, refresh)} & \shortstack{Sequence differs\\(\%)} & \shortstack{Refresh only /\\Static only} & \shortstack{Tail\\score}\\
\midrule
OOD10 & GRN & $4.00,\ 4.00$ & $1.00,\ 4.00$ & 42.64 & 0 / 0 & $16/14$\\
OOD10 & Process & $1.99,\ 1.86$ & $0.00,\ 0.00$ & 27.29 & 7 / 1 & $36/29$\\
OOD10 & Combined & $2.30,\ 2.00$ & $1.00,\ 2.00$ & 27.13 & 0 / 0 & $10/7$\\
OOD10 & Rank meet & $3.95,\ 3.95$ & $1.00,\ 3.95$ & 39.53 & 5 / 0 & $16/10$\\
OOD10 & Borda & $2.00,\ 1.91$ & $1.00,\ 1.91$ & 27.13 & 5 / 0 & $10/6$\\
OOD10 & Rank product & $2.78,\ 2.59$ & $1.00,\ 2.59$ & 30.23 & 0 / 0 & $14/11$\\
\midrule
OOD15 & GRN & $5.10,\ 5.10$ & $1.00,\ 5.10$ & 69.44 & 0 / 0 & $30/19$\\
OOD15 & Process & $2.74,\ 2.47$ & $0.00,\ 0.00$ & 36.76 & 37 / 1 & $61/44$\\
OOD15 & Combined & $3.13,\ 2.64$ & $1.00,\ 2.64$ & 48.84 & 0 / 0 & $16/10$\\
OOD15 & Rank meet & $5.04,\ 5.08$ & $1.00,\ 5.08$ & 68.98 & 0 / 0 & $19/15$\\
OOD15 & Borda & $2.91,\ 2.58$ & $1.00,\ 2.58$ & 47.92 & 0 / 0 & $14/8$\\
OOD15 & Rank product & $3.85,\ 3.44$ & $1.00,\ 3.44$ & 56.94 & 0 / 0 & $18/15$\\
\midrule
OOD20 & GRN & $6.67,\ 6.67$ & $1.00,\ 6.67$ & 76.31 & 5 / 0 & $39/18$\\
OOD20 & Process & $3.27,\ 2.77$ & $0.00,\ 0.00$ & 41.78 & 69 / 0 & $88/58$\\
OOD20 & Combined & $4.06,\ 3.14$ & $1.00,\ 3.14$ & 47.39 & 0 / 0 & $26/12$\\
OOD20 & Rank meet & $6.59,\ 6.64$ & $1.00,\ 6.64$ & 76.31 & 0 / 0 & $25/17$\\
OOD20 & Borda & $3.65,\ 2.96$ & $1.00,\ 2.96$ & 47.91 & 25 / 0 & $34/9$\\
OOD20 & Rank product & $4.87,\ 4.17$ & $1.00,\ 4.17$ & 55.64 & 0 / 0 & $27/15$\\
\bottomrule
\end{tabular}
\end{table}

%% file: grn_cdgs_ordinal_appendix.tex
\paragraph{Ordinal guidance within population-based search.}
Crossing aggregation with statewise refinement tests whether ordinal process and feasibility evidence remain useful within sequence search.  Combined and rank meet each have a Static and a refresh variant, using the same reachable targets, checkpoint, candidate interface, verifier, $5N$ cap, and population parameters specified above.  For the full remaining candidate set, the ordinal sampling utility is
\[
u_{\mathrm{meet}}^s(a)=\operatorname{zscore}_{A_s}\!\left(-\min\{r_T^s(a),r_X^s(a)\}\right),
\]
where the two channels use one-based average weak ranks.  Nonconstant utilities are standardized to unit variance, and tied minimum ranks receive equal utility.  CDGS uses this utility for sequence sampling with the fixed temperature and applies its existing elite guide.  Direct rank meet uses the deterministic ordering defined above.

Static retains both initial raw channels and reconstructs normalization or ranks on each remaining candidate set.  Refresh recomputes both channels in realized and imagined removal states.  Rejection masks the queried pair while preserving the realized state, and imagined removals consume no verifier calls.  Sequence guidance can therefore change the first queried action through scores in imagined successor states.

For each scene, target, and seed, all four variants initialize the same pseudorandom stream.  The population comparisons in Table~\ref{tab:cdgs-main} use method-specific streams.  The crossed comparison retains seeds 0--4 at each scale and averages them within each target before 20,000 paired scene-clustered bootstrap resamples and sign permutations with seed 20260809.  Query scores retain the $5N+1$ failure penalty.  Each policy's scoring demand is counted independently of prediction reuse across methods, targets, and seeds.

\paragraph{Refinement across sampling utilities.}
Table~\ref{tab:cdgs-aggregation-cross} shows refresh gains for both aggregation rules at every scale.  Ordinal guidance reduces mean query scores by $16.66\%$, $12.48\%$, and $19.10\%$ on OOD10, OOD15, and OOD20, respectively, and refreshed policies complete every condition.  All six paired confidence intervals favor refresh.  These results extend the refinement pattern from direct ordinal ranking to imagined sequence selection.  Combined achieves the lowest query scores within this crossed comparison, illustrating the value of choosing an aggregation suited to the available signals.

\begin{table}[htbp]
\centering
\caption{\textbf{Aggregation and refinement within CDGS-style population search.} The four policies share instances, five search seeds, initialization of the random stream, checkpoint, search parameters, verifier, and $5N$ call limit, giving 22,700 runs. Mean query scores include the $5N+1$ failure penalty. $\Delta$ is refresh minus Static with paired scene-clustered 95\% confidence intervals. The reported $p$ values use a two-sided sign permutation test with one shared sign per scene. Completion lists Static first and refresh second. Bold and underlined query scores identify the best and second best among the four policies at each scale.}
\label{tab:cdgs-aggregation-cross}
\footnotesize
\setlength{\tabcolsep}{3pt}
\renewcommand{\arraystretch}{1.05}
\begin{tabular}{llrrlrr}
\toprule
Split & Aggregation & Static $\downarrow$ & Refresh $\downarrow$ & $\Delta$ $[95\%\,\mathrm{CI}]$ & \shortstack{Completion\\(\%)} & $p$\\
\midrule
OOD10 & Combined & 4.116 & \textbf{2.958} & $-1.158\;[-2.105,-0.400]$ & $99.22/100.00$ & 0.00070\\
OOD10 & Rank meet & 4.868 & \underline{4.057} & $-0.811\;[-1.449,-0.239]$ & $100.00/100.00$ & 0.01710\\
\midrule
OOD15 & Combined & 5.769 & \textbf{3.818} & $-1.951\;[-2.538,-1.406]$ & $99.95/100.00$ & 0.00005\\
OOD15 & Rank meet & 5.813 & \underline{5.088} & $-0.725\;[-1.186,-0.301]$ & $100.00/100.00$ & 0.00110\\
\midrule
OOD20 & Combined & 7.505 & \textbf{4.192} & $-3.312\;[-4.168,-2.486]$ & $99.90/100.00$ & 0.00005\\
OOD20 & Rank meet & 7.490 & \underline{6.059} & $-1.431\;[-1.982,-0.964]$ & $99.90/100.00$ & 0.00005\\
\bottomrule
\end{tabular}
\end{table}

Table~\ref{tab:cdgs-aggregation-cost} separates neural scoring from the candidate positions evaluated during sequence sampling.  Refresh uses state-dependent neural predictions to reduce verifier queries and sampled positions under both utilities.  Ordinal guidance explores more scored states than Combined, making the aggregation choice consequential for the balance between neural computation and verifier effort.

\begin{table}[htbp]
\centering
\caption{\textbf{Scoring effort under the crossed aggregation comparison.} Means use all combinations of targets and seeds. Logical GRN forwards count distinct neural inputs when predictions are cached by removal state within each episode. The count includes initial scoring. Static uses one initial forward. Scored positions count candidate positions in imagined sequences. Both metrics list Static first and refresh second, and each policy's requirements are counted independently of prediction reuse across evaluations.}
\label{tab:cdgs-aggregation-cost}
\footnotesize
\setlength{\tabcolsep}{7pt}
\renewcommand{\arraystretch}{1.05}
\begin{tabular}{llrr}
\toprule
Split & Aggregation & \shortstack{Logical GRN forwards\\(Static, refresh)} & \shortstack{Scored positions\\(Static, refresh)}\\
\midrule
OOD10 & Combined & $1.00,\ 12.87$ & $227.56,\ 140.09$\\
OOD10 & Rank meet & $1.00,\ 43.67$ & $305.73,\ 249.14$\\
\midrule
OOD15 & Combined & $1.00,\ 23.19$ & $364.41,\ 213.73$\\
OOD15 & Rank meet & $1.00,\ 88.84$ & $426.13,\ 368.50$\\
\midrule
OOD20 & Combined & $1.00,\ 38.34$ & $531.28,\ 260.45$\\
OOD20 & Rank meet & $1.00,\ 141.02$ & $574.77,\ 452.35$\\
\bottomrule
\end{tabular}
\end{table}

\FloatBarrier

%% file: grn_timing_appendix.tex
\subsubsection{Wall-Clock Verification Value}

\paragraph{Matched timing protocol.}
Wall-clock evaluation measures all six selectors and both variants with the same candidates, scoring operations, annotation verifier, and $5N$ cap as Appendix~\ref{app:grn-search}.  We use the official GRN checkpoint and Panda OOD data \citep{bouhsain2025grn}.  At each scale, the first ten scenes in identifier order are considered, scenes with incomplete movable-object annotations are excluded, and reachable targets are ordered by object identifier.  Retaining at most 100 episodes yields 61 on OOD10, 82 on OOD15, and 100 on OOD20.  All methods share tie seed 0.

Measured execution time includes initial and subsequent scoring, candidate masking, sorting, annotation verification, and state transitions.  Static retains initial raw scores and normalizes or ranks them on remaining candidates; refresh reuses scores after rejections and recomputes them after accepted transitions.  Each method, episode, and repeat starts with an independent score cache.  We average three CUDA-synchronized repeats after model warm-up and rotate method order across repeats.  Model and data loading, target enumeration, warm-up, and auxiliary operations are excluded from timing.

\paragraph{From query savings to execution time.}
Let $t_s,t_r$ be measured execution times and $q_s>q_r$ the mean issued verifier calls for Static and refresh.  An additional verifier latency $\lambda$ gives projected time $t+\lambda q$, so the break-even latency is $\lambda^\star=(t_r-t_s)/(q_s-q_r)$.  This projection uses actual issued calls for every episode, whereas query effectiveness uses the failure-penalized score.  Tables~\ref{tab:timing-full} and~\ref{tab:latency-grid} give the decomposition and projections from these matched measurements.

\paragraph{Scoring overhead and verification cost.}
The measurements identify when verifier savings translate into faster execution.  Combined reaches break-even at $12.55$, $31.26$, and $31.44$ ms per query on OOD10, OOD15, and OOD20, respectively.  Rank meet saves queries while scoring more removal states, giving corresponding thresholds of $35.51$, $86.01$, and $224.50$ ms per query.  Above each threshold, the saved verification time exceeds the added scoring cost (Table~\ref{tab:latency-grid}).  Rank meet refresh and both Combined variants complete every timing episode; rank meet Static completes $98.36\%$ on OOD10 and $100\%$ on both larger scales.

Timing variability is summarized in Table~\ref{tab:timing-variability}.  The standard deviations of mean compute time across repeats are small relative to the scoring overheads in Table~\ref{tab:timing-full}.

\begin{table}[h]
\centering
\caption{\textbf{Variability of wall-clock measurements.} Entries are standard deviations in milliseconds across the three repeats of mean compute time over timing episodes. Populations, scoring operations, and measurement boundaries match Table~\ref{tab:timing-full}.}
\label{tab:timing-variability}
\footnotesize
\renewcommand{\arraystretch}{1.05}
\begin{tabular*}{\textwidth}{@{\extracolsep{\fill}}lrrrr@{}}
\toprule
& \multicolumn{2}{c}{Rank meet} & \multicolumn{2}{c}{Combined}\\
\cmidrule(lr){2-3}\cmidrule(lr){4-5}
Scale & Static & Refresh & Static & Refresh\\
\midrule
OOD10 & 0.249 & 1.167 & 1.299 & 1.025\\
OOD15 & 0.723 & 0.880 & 0.626 & 0.488\\
OOD20 & 1.335 & 2.821 & 0.655 & 1.657\\
\bottomrule
\end{tabular*}
\end{table}

\begin{table}[t]
\centering
\caption{\textbf{Wall-clock decomposition for all matched selectors.} Compute includes initial scoring and is reported in mean milliseconds per episode. Queries are actual issued verifier calls. Paired entries list static first and refresh second. The best and second-best compute means at each scale are bold and underlined. Break-even gives the additional per-query verification latency at which refresh and static have equal projected time.}
\label{tab:timing-full}
\footnotesize
\setlength{\tabcolsep}{3pt}
\renewcommand{\arraystretch}{1.05}
\begin{tabular}{llrrrr}
\toprule
Scale & Selector & \shortstack{Compute (static, refresh)\\(ms/episode) $\downarrow$} & \shortstack{Issued queries\\(static, refresh) $\downarrow$} & Saved queries $\uparrow$ & \shortstack{Break-even\\(ms/query)}\\
\midrule
OOD10 & GRN & 16.578, 54.497 & $5.131,\ 4.820$ & 0.311 & 121.740\\
OOD10 & Process & \textbf{0.214}, \underline{0.237} & $7.967,\ 7.213$ & 0.754 & 0.031\\
OOD10 & Combined & 16.574, 27.681 & $2.951,\ 2.066$ & 0.885 & 12.547\\
OOD10 & Rank meet & 17.637, 56.635 & $5.508,\ 4.410$ & 1.098 & 35.506\\
OOD10 & Borda & 17.052, 27.893 & $2.902,\ 2.492$ & 0.410 & 26.450\\
OOD10 & Rank product & 17.840, 37.514 & $3.869,\ 2.951$ & 0.918 & 21.431\\
\midrule
OOD15 & GRN & 22.904, 99.972 & $7.866,\ 6.537$ & 1.329 & 57.978\\
OOD15 & Process & \textbf{0.382}, \underline{0.431} & $16.195,\ 12.183$ & 4.012 & 0.012\\
OOD15 & Combined & 24.388, 59.335 & $4.171,\ 3.053$ & 1.118 & 31.262\\
OOD15 & Rank meet & 24.763, 103.432 & $6.598,\ 5.683$ & 0.915 & 86.011\\
OOD15 & Borda & 25.246, 61.420 & $4.549,\ 3.744$ & 0.805 & 44.943\\
OOD15 & Rank product & 24.592, 75.495 & $5.366,\ 4.280$ & 1.085 & 46.899\\
\midrule
OOD20 & GRN & 31.540, 172.465 & $9.260,\ 7.610$ & 1.650 & 85.409\\
OOD20 & Process & \textbf{0.473}, \underline{0.514} & $15.060,\ 10.970$ & 4.090 & 0.010\\
OOD20 & Combined & 32.994, 80.463 & $4.810,\ 3.300$ & 1.510 & 31.436\\
OOD20 & Rank meet & 34.273, 173.465 & $7.870,\ 7.250$ & 0.620 & 224.503\\
OOD20 & Borda & 32.457, 72.484 & $4.580,\ 3.130$ & 1.450 & 27.605\\
OOD20 & Rank product & 33.346, 102.912 & $5.740,\ 4.480$ & 1.260 & 55.211\\
\bottomrule
\end{tabular}
\end{table}

\begin{table}[h]
\centering
\caption{\textbf{Projected end-to-end latency difference: refresh minus static.} Entries are milliseconds per episode; negative values favor refresh. Projection adds the stated additional latency to every actual verifier call. At each scale and latency, the best and second-best differences among selectors are bold and underlined.}
\label{tab:latency-grid}
\footnotesize
\setlength{\tabcolsep}{2.3pt}
\renewcommand{\arraystretch}{1.05}
\begin{tabular}{llrrrrrr}
\toprule
Scale & Selector & 1 ms/query & 10 ms/query & 50 ms/query & 100 ms/query & 500 ms/query & 1000 ms/query\\
\midrule
OOD10 & GRN & 37.61 & 34.80 & 22.35 & 6.77 & -117.82 & -273.56\\
OOD10 & Process & \textbf{-0.73} & \textbf{-7.52} & \textbf{-37.68} & \underline{-75.39} & -377.03 & -754.08\\
OOD10 & Combined & \underline{10.22} & \underline{2.25} & \underline{-33.16} & \textbf{-77.42} & -431.52 & -874.14\\
OOD10 & Rank meet & 37.90 & 28.01 & -15.92 & -70.84 & \textbf{-510.18} & \textbf{-1059.36}\\
OOD10 & Borda & 10.43 & 6.74 & -9.65 & -30.14 & -194.08 & -399.00\\
OOD10 & Rank product & 18.76 & 10.49 & -26.23 & -72.13 & \underline{-439.34} & \underline{-898.36}\\
\midrule
OOD15 & GRN & 75.74 & 63.78 & 10.60 & -55.86 & \underline{-587.57} & \underline{-1252.20}\\
OOD15 & Process & \textbf{-3.96} & \textbf{-40.07} & \textbf{-200.56} & \textbf{-401.17} & \textbf{-2006.05} & \textbf{-4012.15}\\
OOD15 & Combined & \underline{33.83} & \underline{23.77} & \underline{-20.95} & \underline{-76.84} & -524.00 & -1082.94\\
OOD15 & Rank meet & 77.75 & 69.52 & 32.94 & -12.79 & -378.65 & -835.97\\
OOD15 & Borda & 35.37 & 28.13 & -4.07 & -44.31 & -366.27 & -768.70\\
OOD15 & Rank product & 49.82 & 40.05 & -3.37 & -57.63 & -491.78 & -1034.46\\
\midrule
OOD20 & GRN & 139.28 & 124.43 & 58.43 & -24.07 & -684.07 & \underline{-1509.07}\\
OOD20 & Process & \textbf{-4.05} & \textbf{-40.86} & \textbf{-204.46} & \textbf{-408.96} & \textbf{-2044.96} & \textbf{-4089.96}\\
OOD20 & Combined & 45.96 & 32.37 & -28.03 & -103.53 & \underline{-707.53} & -1462.53\\
OOD20 & Rank meet & 138.57 & 132.99 & 108.19 & 77.19 & -170.81 & -480.81\\
OOD20 & Borda & \underline{38.58} & \underline{25.53} & \underline{-32.47} & \underline{-104.97} & -684.97 & -1409.97\\
OOD20 & Rank product & 68.31 & 56.97 & 6.57 & -56.43 & -560.43 & -1190.43\\
\bottomrule
\end{tabular}
\end{table}